%% file: main.tex
\documentclass[10pt,twocolumn,letterpaper]{article}

\usepackage[pagenumbers]{cvpr} % To force page numbers, e.g. for an arXiv version


\definecolor{cvprblue}{rgb}{0.21,0.49,0.74}
\usepackage[pagebackref,breaklinks,colorlinks,allcolors=cvprblue]{hyperref}
\usepackage{graphicx}

\def\paperID{*****} % *** Enter the Paper ID here
\def\confName{CVPR}
\def\confYear{2025}

\title{Naturally Computed Scale Invariance in the Residual Stream of ResNet18}

\author{Andr\'e Longon\\
Independent\\
{\tt\small ajl321@hotmail.com}
}

\begin{document}
\maketitle
\input{sec/0_abstract}    
\input{sec/1_intro}
\input{sec/2_methods}
\input{sec/3_criteria}
\input{sec/4_ablation}
\input{sec/5_discussion}
\input{sec/6_conclusion}
{
    \small
    \bibliographystyle{ieeenat_fullname}
    \bibliography{main}
}

\input{sec/X_suppl}

\end{document}

%% file: sec/0_abstract.tex
\begin{abstract}
An important capacity in visual object recognition is invariance to image-altering variables which leave the identity of objects unchanged, such as lighting, rotation, and scale.  How do neural networks achieve this?  Prior mechanistic interpretability research has illuminated some invariance-building circuitry in InceptionV1, but the results are limited and networks with different architectures have remained largely unexplored.  This work investigates ResNet18 with a particular focus on its residual stream, an architectural component which InceptionV1 lacks.  We observe that many convolutional channels in intermediate blocks exhibit scale invariant properties, computed by the element-wise residual summation of scale equivariant representations: the block input's smaller-scale copy with the block pre-sum output's larger-scale copy.  Through subsequent ablation experiments, we attempt to causally link these neural properties with scale-robust object recognition behavior.  Our tentative findings suggest how the residual stream computes scale invariance and its possible role in behavior.  Code is available at: \url{https://github.com/cest-andre/residual-stream-interp}
\end{abstract}

%% file: sec/1_intro.tex
\section{Introduction}
\label{sec:intro}

Deep neural networks (DNNs) and brains learn to perform sophisticated tasks but leave us ignorant of how they ultimately solve them.  One such task is visual object recognition, which maps an image onto an object category.  A critical property of object recognition is invariance: to recognize objects across arbitrary positions, angles, scales (from depth or size), lighting, and backgrounds.  How do artificial and biological neural networks accomplish this?  Invariant object recognition is a central focus of study in visual neuroscience \cite{dicarlo2012does}, but the complexities of the brain have rendered it difficult to elucidate the underlying neural algorithms, especially in deeper regions of the ventral stream.  Alternatively, as DNNs are the leading models of neural response predictions in visual processing \cite{yamins2014performance, khaligh2014deep, Schrimpf407007, ratan2021computational}, reverse-engineering these models may help uncover the mechanisms from which invariant object recognition emerges in the brain.

In the digital realm, the field of mechanistic interpretability has progressed in elucidating circuitry of object recognition DNNs \cite{olah2020zoom}.  InceptionV1 \cite{Szegedy_2015_CVPR} is the most well-studied DNN where many features have been categorized \cite{olah2020an}, including detectors for curves \cite{cammarata2020curve, gorton2024the} and high-low frequency boundaries \cite{schubert2021high-low}.  Most relevant to this study, Olah \etal \cite{olah2020naturally} discovered within InceptionV1 equivariances for hue, rotation, and scale, as well as some circuitry that build invariances.

To expand the mechanistic interpretation of invariant object recognition in DNNs, we zoom into the \textit{residual stream} of ResNet18 \cite{He_2016_CVPR}.  The residual stream, which InceptionV1 lacks, is an architectural mechanism that permits features to bypass layers of processing via their summation to the output of a downstream layer.  We selected ResNet18 as it is the smallest member of the ResNet family, most manageable to perform a network-scale study of its stream.  We present the following results:

\begin{enumerate}
    \item With simple quantitative criteria, we find residual stream channels which exhibit properties of scale invariance.  The input channel of a block possesses a smaller scale version of a feature, the same channel at the block's pre-sum output possesses the larger scale, and these are summed together to produce the final scale invariant output.  This pattern is especially pronounced in intermediate blocks.

    \item When these channels are ablated, they cause more damage to scale-robust object recognition performance compared to other random channels.  This is evidence of a causal link from scale invariant neurons to scale invariant behavior.
\end{enumerate}

Furthermore, this work suggests a function for so-called ``bypass connections'' in biological visual systems (macaque \cite{nakamura1993modular} and mouse \cite{d2022hierarchical}), an anatomical analogue to the residual stream.  The hypothesis that bypass connections may compute scale invariance is perhaps worth exploring for visual neuroscience and to further establish universality between artificial and biological neural networks.

%% file: sec/2_methods.tex
\begin{figure*}[h]
    \centering
    \begin{subfigure}{.16\linewidth}
        \centering
        \includegraphics[width=\linewidth]{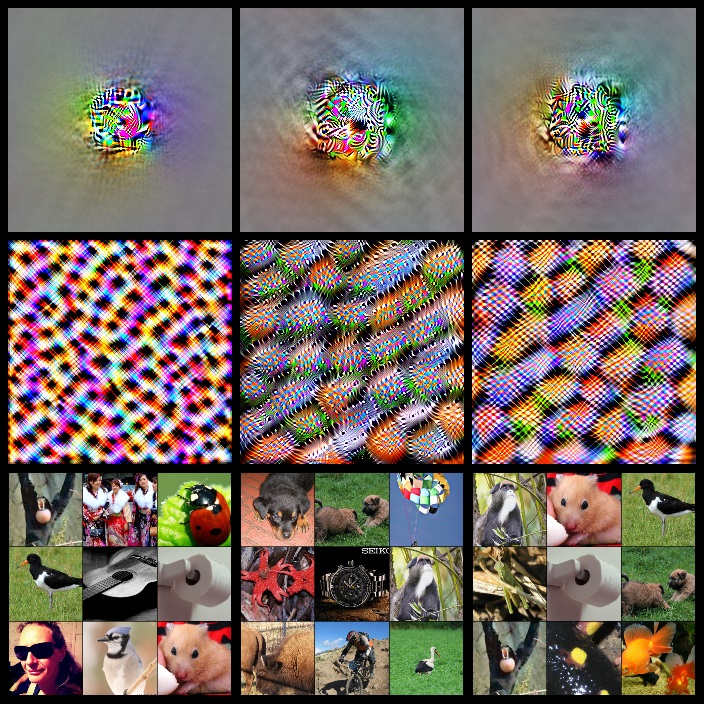}
        \textbf{2.1}: channel 15
        \vspace{0.15\linewidth}
    \end{subfigure}
    \begin{subfigure}{.16\linewidth}
        \centering
        \includegraphics[width=\linewidth]{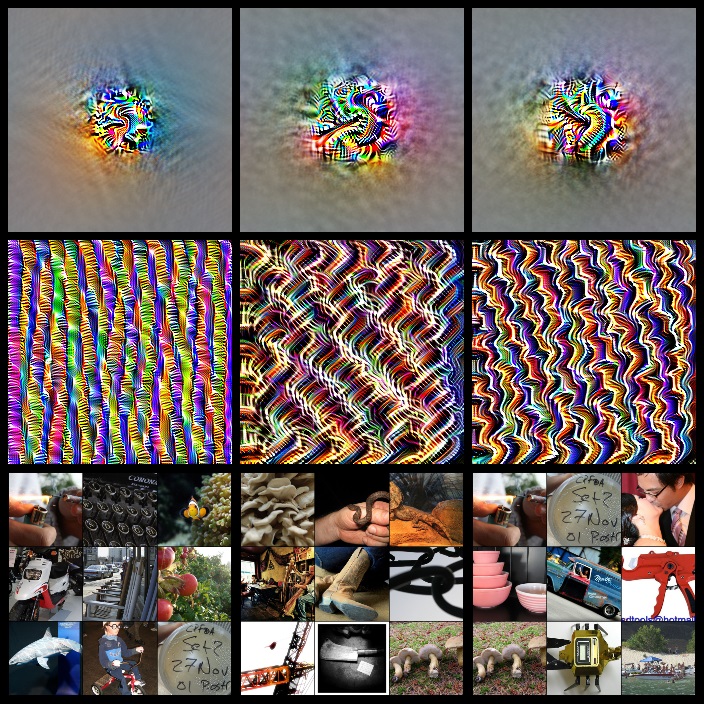}
        21
        \vspace{0.15\linewidth}
    \end{subfigure}
    \begin{subfigure}{.16\linewidth}
        \centering
        \includegraphics[width=\linewidth]{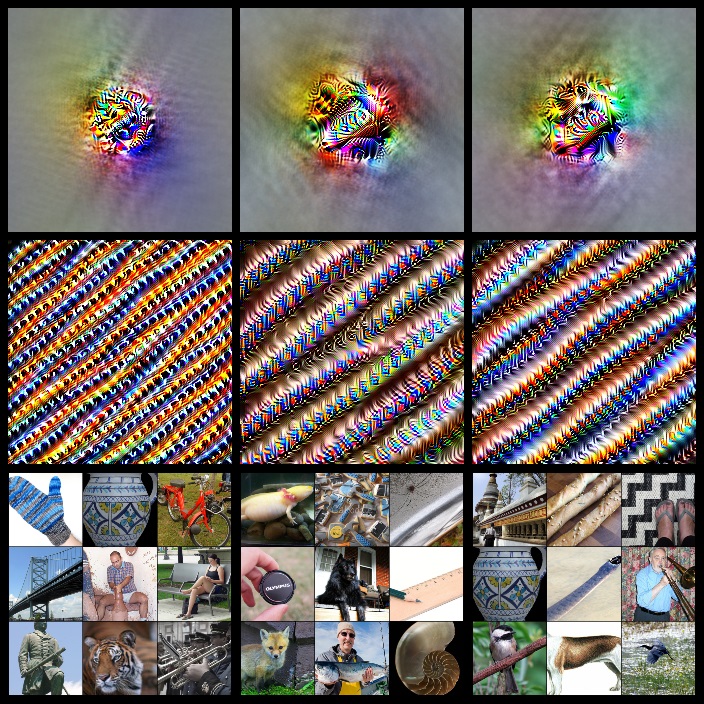}
        48
        \vspace{0.15\linewidth}
    \end{subfigure}
    \begin{subfigure}{.16\linewidth}
        \centering
        \includegraphics[width=\linewidth]{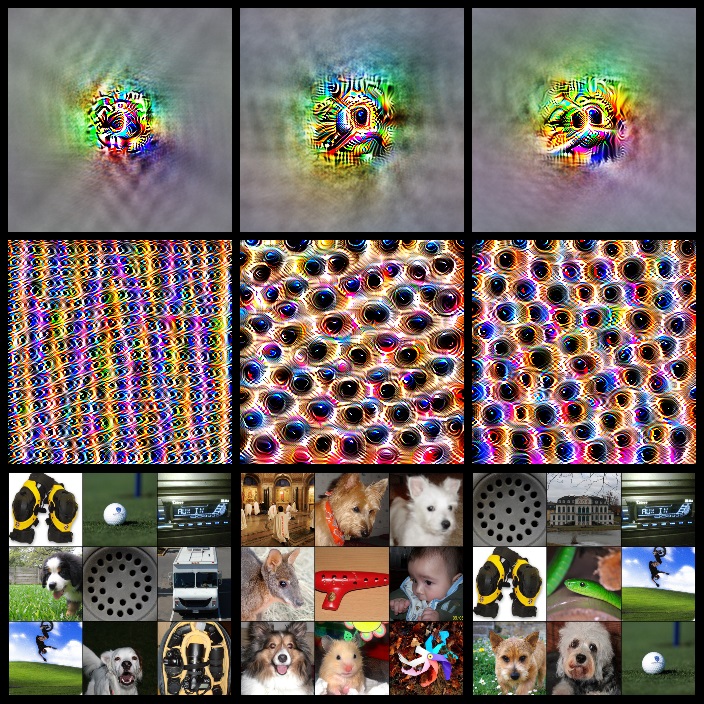}
        68
        \vspace{0.15\linewidth}
    \end{subfigure}
    \begin{subfigure}{.16\linewidth}
        \centering
        \includegraphics[width=\linewidth]{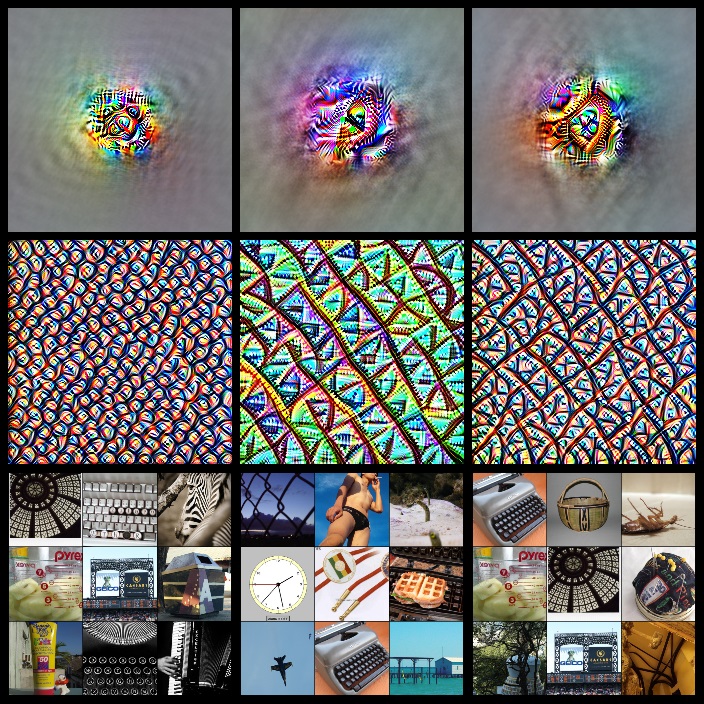}
        88
        \vspace{0.15\linewidth}
    \end{subfigure}
    \begin{subfigure}{.16\linewidth}
        \centering
        \includegraphics[width=\linewidth]{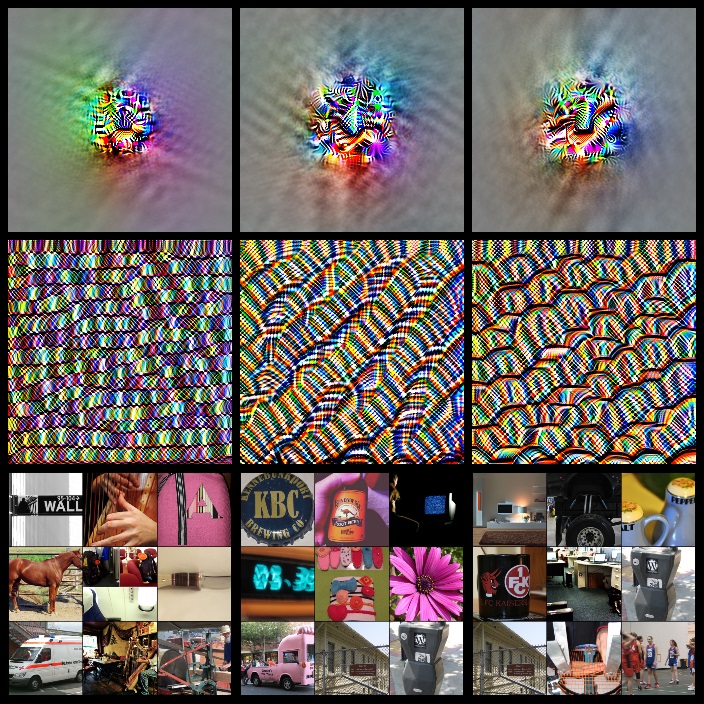}
        108
        \vspace{0.15\linewidth}
    \end{subfigure}
    
    \begin{subfigure}{.16\linewidth}
        \centering
        \includegraphics[width=\linewidth]{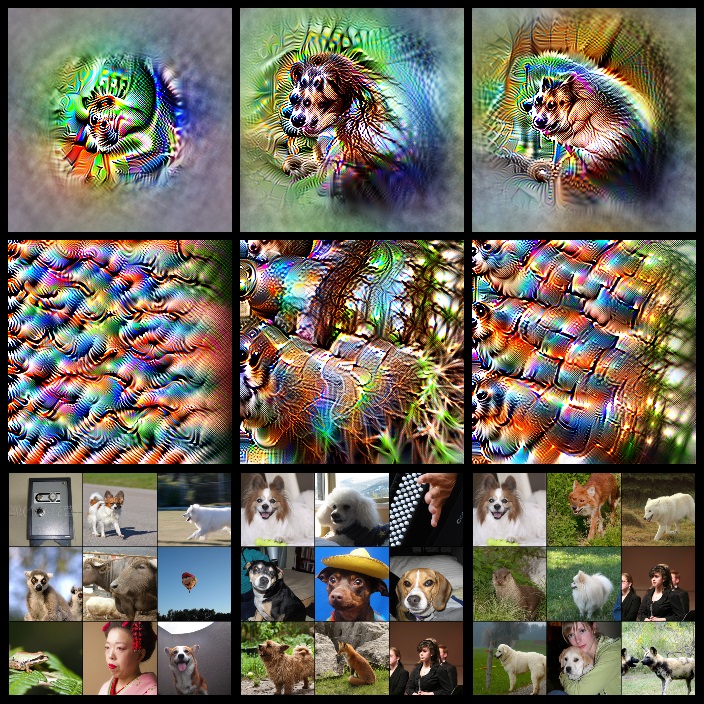}
        \textbf{3.1}: channel 22
    \end{subfigure}
    \begin{subfigure}{.16\linewidth}
        \centering
        \includegraphics[width=\linewidth]{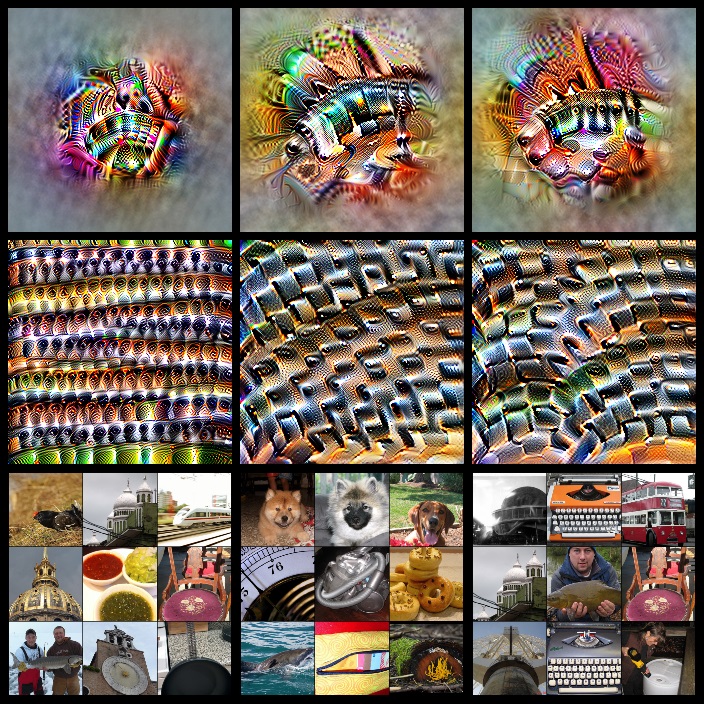}
        29
    \end{subfigure}
    \begin{subfigure}{.16\linewidth}
        \centering
        \includegraphics[width=\linewidth]{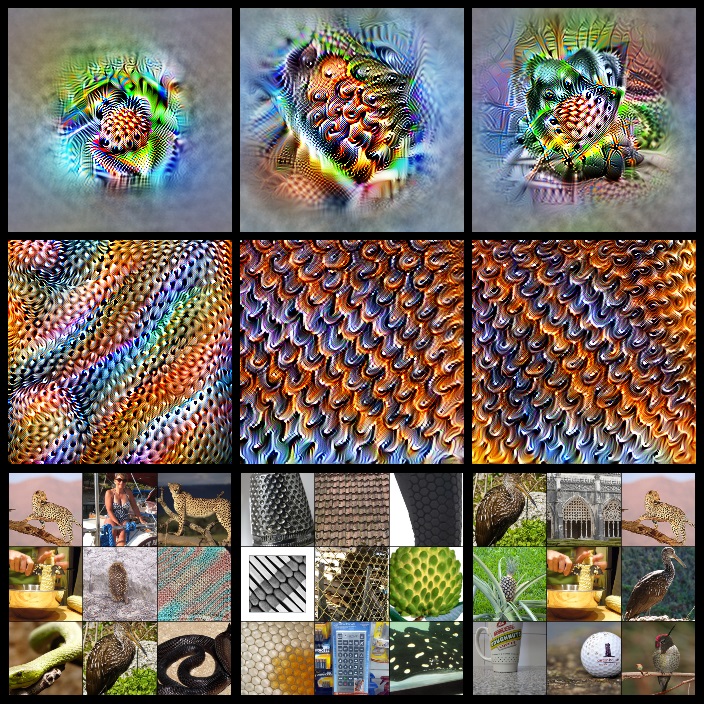}
        80
    \end{subfigure}
    \begin{subfigure}{.16\linewidth}
        \centering
        \includegraphics[width=\linewidth]{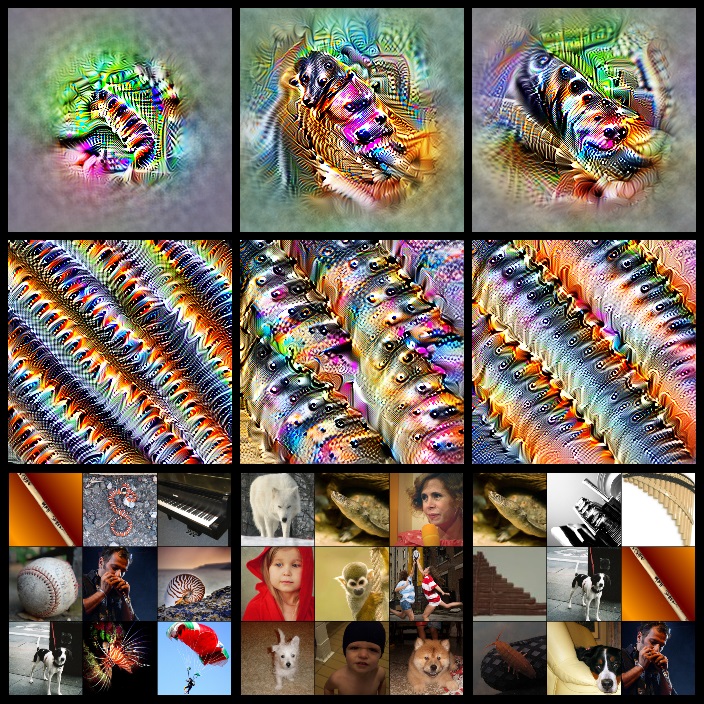}
        113
    \end{subfigure}
    \begin{subfigure}{.16\linewidth}
        \centering
        \includegraphics[width=\linewidth]{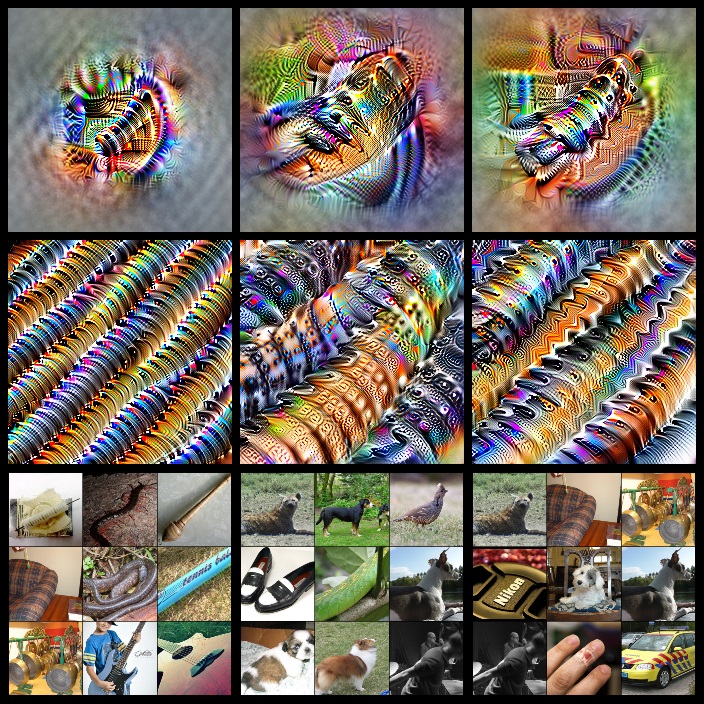}
        178
    \end{subfigure}
    \begin{subfigure}{.16\linewidth}
        \centering
        \includegraphics[width=\linewidth]{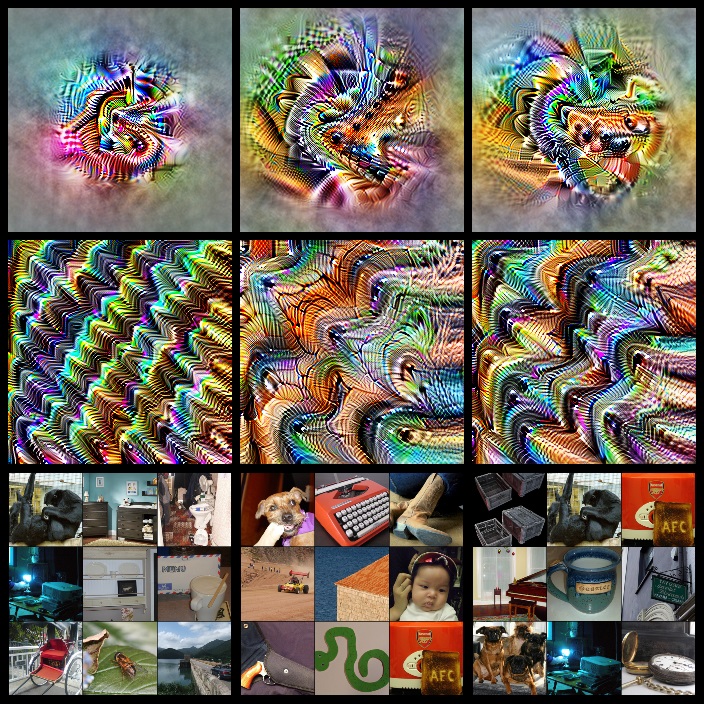}
        215
    \end{subfigure}
    
    \caption{Grids of maximally exciting images for exemplary scale invariant channels in block 2.1 (top row) and 3.1 (bottom row).  All channels are zero-indexed.  Each 3x3 grid is for a single channel across the block's target layers and is organized as follows.  Columns from left to right:  block channel at input, pre-sum, and post-sum layers.  Rows from top to bottom:  FZs optimized for the channel's center neuron, FZs for the entire channel, and the top 9 center-neuron activating natural images from the ImageNet validation set \cite{5206848}.}
    \label{fig:1}
\end{figure*}

\section{Methods}
\label{sec:methods}

We use the PyTorch \cite{NEURIPS2019_bdbca288} torchvision implementation of ResNet18 with the default ImageNet-trained \cite{5206848} weights, \verb|IMAGENET1K_V1|.  Neuron activations are obtained via the THINGSvision library \cite{Muttenthaler_2021}. 

\subsection{Residual stream architecture}

This work is only concerned with the activations from three layers for a given ResNet block.  We name the layers \(\boldsymbol{In}\), \(\boldsymbol{Pre}\), and \(\boldsymbol{Post}\) for the block's input, pre-sum output, and post-sum output respectively (the block index is not notated as we analyze them in isolation).  \(\boldsymbol{Pre}\) is the block's final batch normalization.  \(\boldsymbol{In}\) is the layer whose output bypasses the main layers of the block and added to the output of \(\boldsymbol{Pre}\) (either previous block's output or downsample module output).  \(\boldsymbol{Post}\) is the sum of outputs from \(\boldsymbol{In}\) and \(\boldsymbol{Pre}\) followed by a ReLU.  We denote the center-neuron activation of layer \(\boldsymbol{In}\) in channel \(c\) to image \(X\) as \(\boldsymbol{In}_c(X)\) (equivalent notation for \(\boldsymbol{Pre}\) and \(\boldsymbol{Post}\)).

\subsection{Feature visualization}

Our results rely on feature visualization, an optimization algorithm used to generate images which maximally activate a target channel (or an individual neuron) in a given layer.  Such images are henceforth referred to as feature visualizations (FZs).  We use the version of feature visualization described by \cite{olah2017feature} and implemented in the Lucent library\footnote{\url{https://github.com/greentfrapp/lucent}} for PyTorch.  We use near-identical regularizations as prescribed by \cite{olah2017feature} for improved interpretability, with altered jitter values described in the appendix (see Section \ref{fz_reg}).  We use a channel's center-neuron optimized FZs for all subsequent analyses.

In a given block of the residual stream, for a given channel \(c\), we obtain the center-neuron FZs for the three target layers, which we denote as \(\hat{X}_{In_c}\), \(\hat{X}_{Pre_c}\), and \(\hat{X}_{Post_c}\).  Thus, \(\boldsymbol{In}_c(\hat{X}_{In_c})\) is the center-neuron activation elicited by its own FZ for channel \(c\) of a given block's input layer.

%% file: sec/3_criteria.tex
\section{Criteria reveal scale invariant channels}
\label{sec:criteria}

\begin{figure*}[h]
    \centering
    \begin{subfigure}{.45\linewidth}
        \centering
        \includegraphics[width=\linewidth]{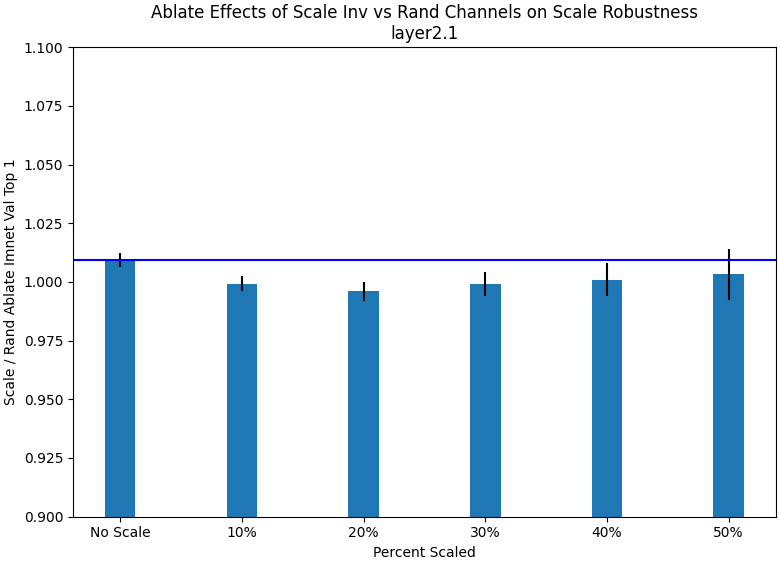}
    \end{subfigure}
    \hfill
    \begin{subfigure}{.44\linewidth}
        \centering
        \includegraphics[width=\linewidth]{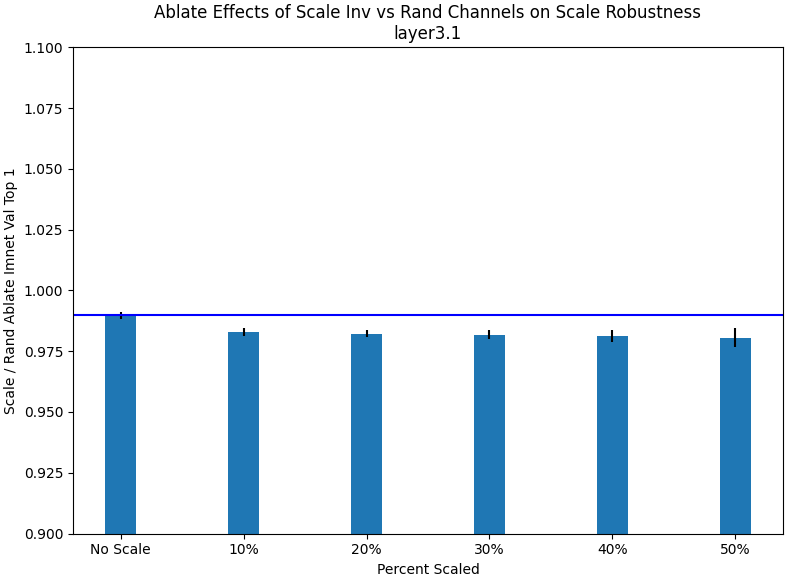}
    \end{subfigure}
    \caption{Scale-robust object recognition degradation from ablating scale invariant channels versus random non-invariant channels in block 2.1 (left) and 3.1 (right).  The y-axis shows the mean ratios of top-1 ImageNet validation accuracy \cite{5206848} between the two ablation conditions at a given scale transformation percentage (applied to all images), with black bars showing the standard error across the random trials.  The blue line is the ratio when no scale transform is applied.  The scale-transformed ratios are below the blue line, meaning scale invariant channel ablation did disproportionately more damage to accuracy during scale-robust object recognition.}
    \label{fig:2}
\end{figure*}

We assume that if a particular post-sum output channel \(\boldsymbol{Post}_c\) at a given block is to be scale invariant, then \(\boldsymbol{In}_c\) will represent the smaller-scale copy, and \(\boldsymbol{Pre}_c\) will represent the larger-scale copy.  This is a reasonable assumption as the receptive field of a neuron in \(\boldsymbol{In}_c\) is smaller than the one in \(\boldsymbol{Pre}_c\) (or equal in deep enough blocks).  We then determine if \(\boldsymbol{Post}_c\) exhibits scale invariance by checking if both of the following conditions are true:
\begin{gather} \label{scale_eq:1}
    \textrm{ReLU}(\boldsymbol{Pre}_c(\hat{X}_{In_c})) < \boldsymbol{Pre}_c(S(\hat{X}_{In_c}))\\
\label{scale_eq:2}
    \frac{2}{3} < \frac{\boldsymbol{Post}_c(\hat{X}_{In_c})}{\boldsymbol{Post}_c(\hat{X}_{Pre_c})} < \frac{3}{2}
\end{gather}
where \(S(X)\) denotes a scale transformation of an image which first center-crops to 112 pixels (half the spatial resolution), then resizes back to 224 pixels using bilinear interpolation.

Both these criteria ensure that increasing the scale of an input channel's FZ will increase the activation of the pre-sum channel in the positive domain (\cref{scale_eq:1}) \textit{and} that the post-sum channel activation is invariant (with some tolerance) to both summands' FZs (\cref{scale_eq:2}).

\paragraph{Results}

We probed all the channels from blocks 1.1 to 4.1 and four blocks contained at least one criteria-passing channel: 1.1 with six channels (\(\sim9.4\%\) of all channels in block), 2.0 with two (\(\sim1.5\%\)), 2.1 with 23 (\(\sim18\%\)), and 3.1 with 46 channels (curiously, \(\sim18\%\) as well).  We note that results may vary due to the stochastic nature of feature visualization.  In \cref{fig:1}, we show exemplary channels which pass the criteria in blocks 2.1 (top row) and 3.1 (bottom row).  The remainder of all passing channels are displayed in the appendix \cref{fig:A1} and \cref{fig:A2}.  

Subjectively, it appears that the criteria unearth channels with compelling scale invariant properties, especially vivid in block 3.1.  We see a variety of features: oriented cylinders (channels 29, 113, 178), a spotted texture (80), and a complex curve (215), where each channel of which has a smaller scale version preferred at the block input, and a larger scale preferred at the pre-sum output.  We also note that the center-neuron top-activating ImageNet \cite{5206848} validation images for a channel contains no overlap between the input and pre-sum layers, but the post-sum layer often contains a mixture of them in addition to novel images.  This is expected if the post-sum channels are indeed selective for the same feature across a range of scales, but we leave investigations with natural images for future work.

%% file: sec/4_ablation.tex
\section{Ablation effects on object recognition}

We next attempt to connect scale invariant \textit{neural} properties with ``scale-robust''  \textit{behavior} during object recognition.  We operationalize scale robustness by recording ImageNet validation top-1 classification accuracy on an ablated ResNet18 while a scale transformation is applied on all images.  Our hypothesis is that ablating the scale invariant channels of a block will disproportionately damage scale-robust accuracy more than when ablating non-scale invariant random channels.

The transformation first applies a center-crop to perform the scaling.  The degree of scale is expressed as a percentage of the input pixel resolution (256), \eg, 10\% center-crops to 231 pixels (\(256-\lfloor256*0.1\rfloor\)).  Then a bilinear interpolation resize to the final pixel resolution of 224 is performed.  We sweep through a range of five scale percentages, 10\% through 50\% at 10\% steps, and each trial applies the same percentage transform throughout the entire validation set.

For a given scale percentage, we measure accuracy for two ablation conditions: (1) all criteria-passing channels in a block are ablated, (2) the same number of non-passing channels are randomly selected and ablated.  We perform mean ablation of the post-sum activations: each post-sum channel's center-neuron activations across the validation set are obtained and the mean is taken.  Then for each batch during the classification trial, we overwrite the selected channels' post-sum activations to their respective center means across all spatial positions.  We only perform the experiments on blocks 2.1 and 3.1, where the proportion of criteria-passing channels are greatest, to produce larger ablation effects.

It is possible that scale invariant channels may produce more damage due to their features having greater object recognition usefulness aside from building scale invariance (\eg, more detectors for dog-related features).  To control for this, we screen for random channels that \textit{do more accuracy damage than the scale invariance channels when no scale transform is applied}, then ablate these same channels across the different scale percentages.  This was achieved for block 2.1, but it was rare to find such channels in 3.1, so this constraint was relaxed such that random ablation accuracy can be no more than 1\% greater compared to the scale invariant ablation accuracy in the no-scale condition.

A total of 10 trials for each block were performed for the random channel ablation condition.  We present the accuracies at a given scale transformation as a ratio between the two ablation conditions, averaged over all trials:

\begin{equation}
    \frac{1}{N} \sum_{i=1}^{N} \frac{scale\_ablate\_acc}{rand\_ablate\_accs_i}
\end{equation}
where \(N\) is the number of trials.

\paragraph{Results}

\cref{fig:2} presents our findings for block 2.1 (left) and 3.1 (right).  It is critical for our hypothesis to compare accuracy ratios in scale-transformed accuracies relative to the no-scale ratio (blue line in \cref{fig:2}).  In the null hypothesis, all channels have similar ablation effects on scale-transformed versus no-scale trials, so we would expect the ratios for two sets of channels across the scale transforms to be equal to the no-scale ratio.  \textit{However, we instead see that the ratios for the scale-transformed trials are consistently lower than the no-scale trials.}  In other words, ablating the scale invariant channels during scale-transformed trials does more accuracy damage relative to the no-scale trials than would be expected in the null hypothesis.  While the effects are small, they are consistent across scale percentages and different random channel samples.  We take this as evidence that the scale invariant representations built by the residual stream support scale-robust object recognition behavior.

%% file: sec/5_discussion.tex
\section{Discussion}

\paragraph{Limitations} The criteria for determining scale invariance is only via a channel's response to the FZs.  To test the generalization of these nascent results, future work must incorporate responses to a wider range of natural images.  More definitive criteria should be developed, such as the circuit-level analysis used to reveal equivariances in InceptionV1 \cite{olah2020naturally}.  If scaled versions of the same feature exist in the input and pre-sum layers of a block, they must also have correspondingly scaled circuitry.

Our ablation results are also quite limited as they show only a small effect, leaving open the possibility that scale-robust accuracies may rely on additional mechanisms beyond the residual stream.  Future work may want to train models with and without a residual stream from scratch to measure the degree of scale-robust behavior that resides in a streamless, yet otherwise equivalent, architecture. 

\paragraph{Future Directions} Our work focused on individual neurons/channels, the ``bases'' of a given layer's activation space.  It is possible there are more scale invariant features represented by the layer's neural population, but obscured by superposition \cite{elhage2022superposition}.  A promising approach to reveal such features is the group crosscoder \cite{gorton2024group}, a dictionary learning technique which extracts equivariant features from superposition in a given layer.  Perhaps this tool can be adapted to find additional scale equivariant features summed in the residual stream to form invariance.

We also leave open a more thorough search for scale invariant properties across models, such as the remainder of the ResNet family.  We consider recurrent neural networks such as CORNet-S \cite{kubilius2019brain} especially intriguing.  As weights are shared across recurrent timesteps, this allows for the detection of the identical feature across different scales, and its bypass connection can build the invariance.

As mentioned, we hope these findings inspire visual neuroscience to investigate if building scale invariance is a function of bypass connections, for it is known that deeper ventral stream neurons have some degree of scale invariance \cite{ito1995size}.  This would work by directly connecting upstream regions which represent smaller-scale features to downstream regions which represent the larger scales.  A prediction would be that scale invariant deep neurons would respond faster to smaller scales than larger ones due to bypassing the intermediate areas.  Progress in this direction may find universality of visual processing between artificial and biological neural networks, mechanistically explaining \textit{why} DNNs perform so well at predicting neural responses to visual stimuli in the ventral stream.

%% file: sec/6_conclusion.tex
\section{Conclusion}

This work presents tentative evidence for scale invariant representations computed via the residual stream in ResNet18.  We use feature visualization to test scale invariant properties at the neural level, and perform ablation experiments to test their causal role in scale-robust object recognition behavior.  Our results find that intermediate blocks contain a large degree of channels with scale invariant properties, and ablating these channels produces disproportionate object recognition damage under a range of scale transformations. 

As invariance to scale is a critical ingredient for invariant object recognition, these findings are crucial to further the mission of mechanistically understanding this sophisticated cognitive ability.  In addition, our work offers exciting possibilities to discover analogous mechanisms in biological neural networks, which can establish artificial-biological universality and explain why DNNs possess such high predictive power in visual neuroscience.

%% file: sec/X_suppl.tex
\clearpage
\setcounter{page}{1}
\maketitlesupplementary

\section{Appendix}

\subsection{Feature Visualization Regularization} \label{fz_reg}

We use an identical regularization method as described in \cite{olah2017feature}, except we use different levels of jitter dependent on the depth of the layer (we still use half the initial jitter value for the second jitter transform).  This is done due to the unreliability high jitter has at producing a high activation on the center neuron once optimization has completed.  We sacrificed some interpretability that jitter offers for more accurate activation values.

We omitted jitter for the entire 1.1 and 2.0 blocks, we used jitter\(=\)4 for 2.1 \(\boldsymbol{Pre}\) and the entire 3.0 block, then used the default jitter\(=\)16 value for the remainder of the blocks.

\subsection{Visualization grids for remaining scale invariant channels} \label{grids}

\begin{figure*}[t]
    \centering
    \begin{subfigure}{.16\linewidth}
        \centering
        \includegraphics[width=\linewidth]{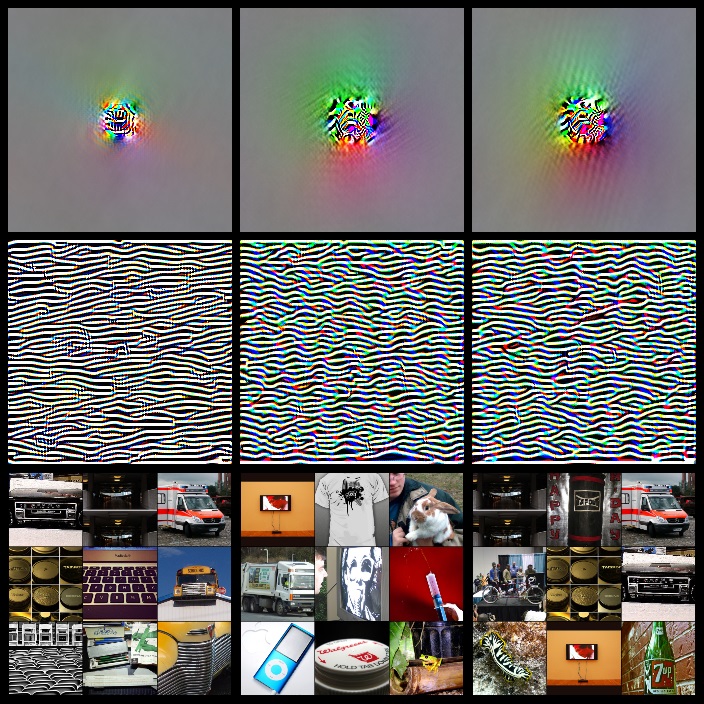}
        \textbf{1.1}: channel 10
        \vspace{0.15\linewidth}
    \end{subfigure}
    \begin{subfigure}{.16\linewidth}
        \centering
        \includegraphics[width=\linewidth]{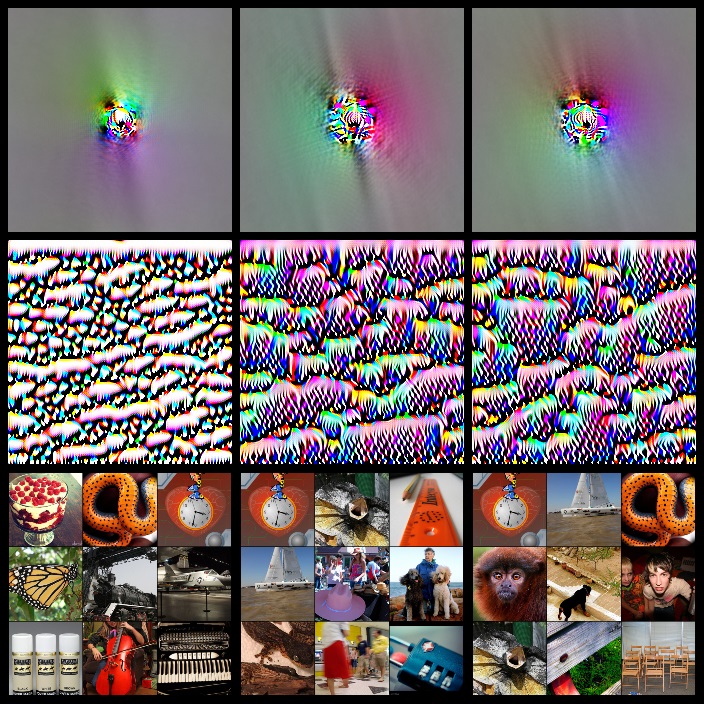}
        19
        \vspace{0.15\linewidth}
    \end{subfigure}
    \begin{subfigure}{.16\linewidth}
        \centering
        \includegraphics[width=\linewidth]{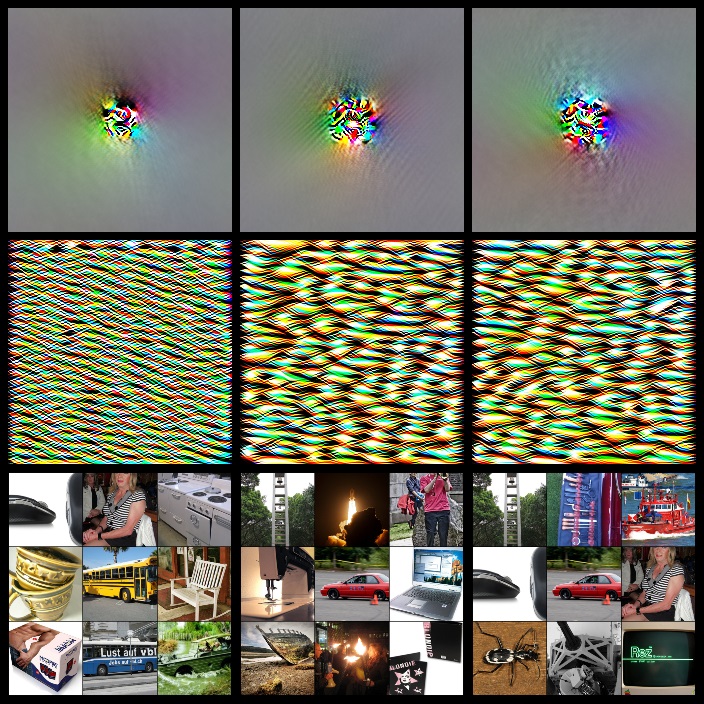}
        21
        \vspace{0.15\linewidth}
    \end{subfigure}
    \begin{subfigure}{.16\linewidth}
        \centering
        \includegraphics[width=\linewidth]{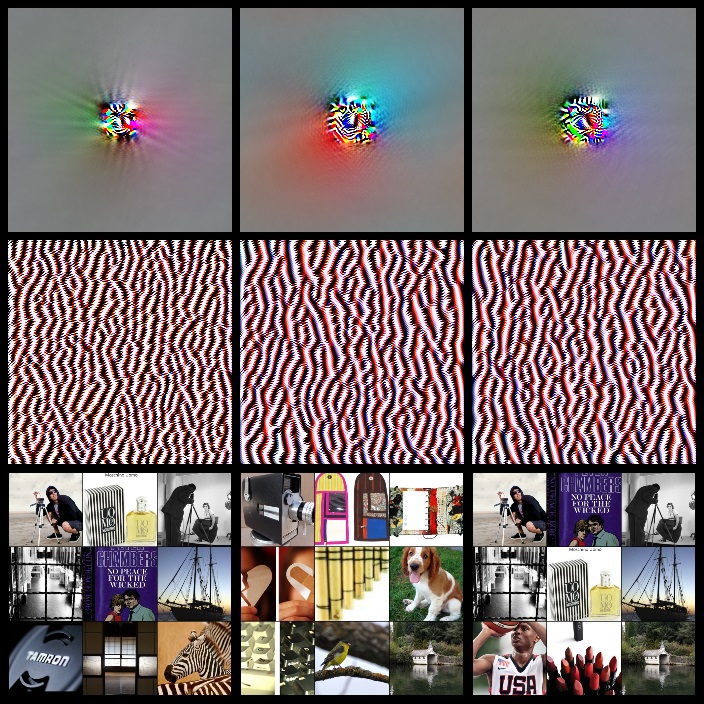}
        25
        \vspace{0.15\linewidth}
    \end{subfigure}
    \begin{subfigure}{.16\linewidth}
        \centering
        \includegraphics[width=\linewidth]{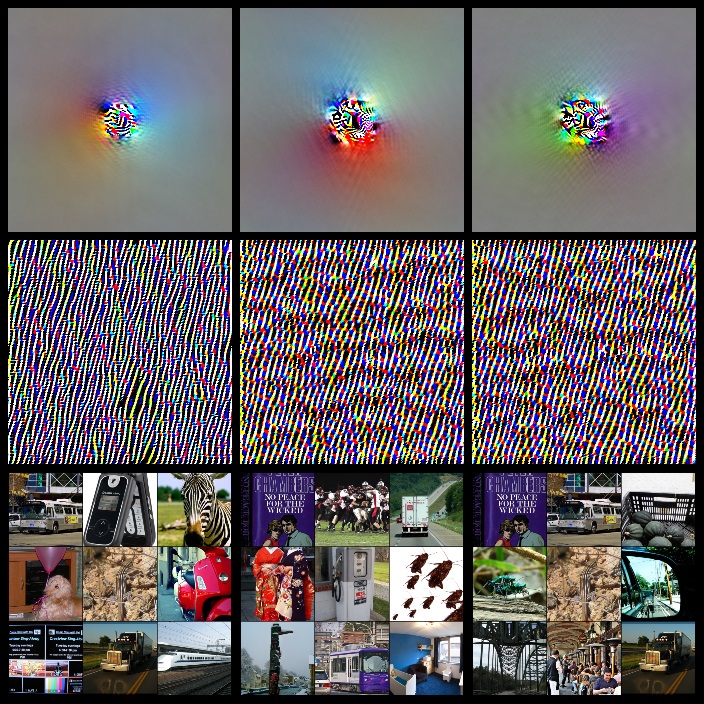}
        44
        \vspace{0.15\linewidth}
    \end{subfigure}
    \begin{subfigure}{.16\linewidth}
        \centering
        \includegraphics[width=\linewidth]{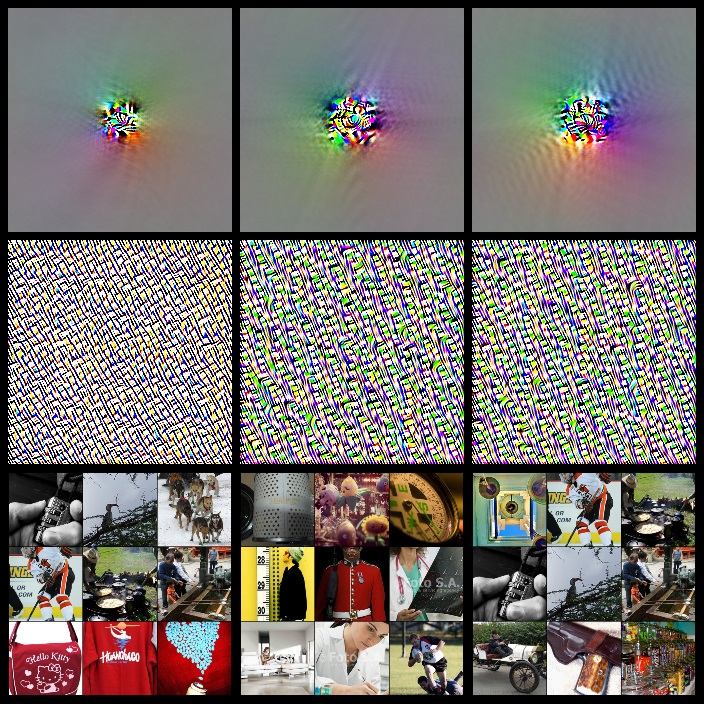}
        50
        \vspace{0.15\linewidth}
    \end{subfigure}
    
    \begin{subfigure}{.16\linewidth}
        \centering
        \includegraphics[width=\linewidth]{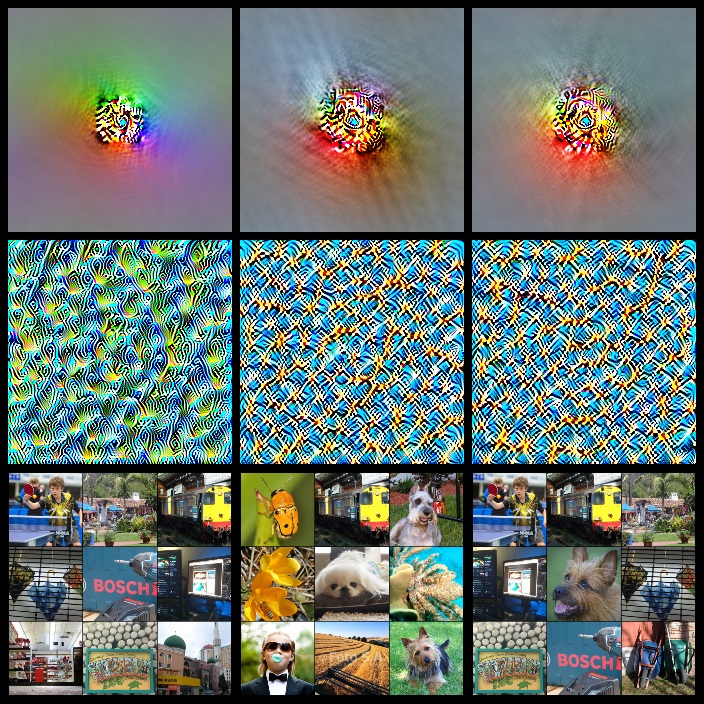}
        \textbf{2.0}: channel 50
        \vspace{0.15\linewidth}
    \end{subfigure}
    \begin{subfigure}{.16\linewidth}
        \centering
        \includegraphics[width=\linewidth]{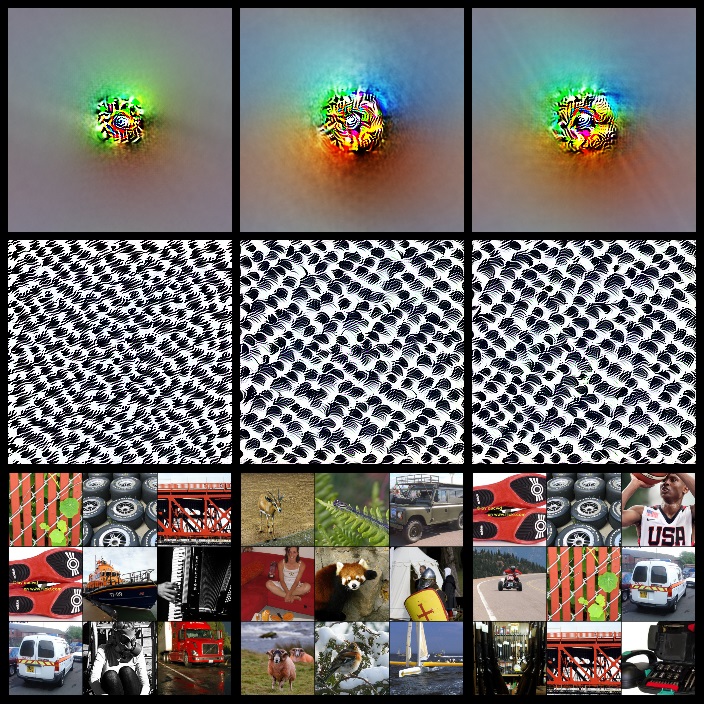}
        53
        \vspace{0.15\linewidth}
    \end{subfigure}

    \begin{subfigure}{.16\linewidth}
        \centering
        \includegraphics[width=\linewidth]{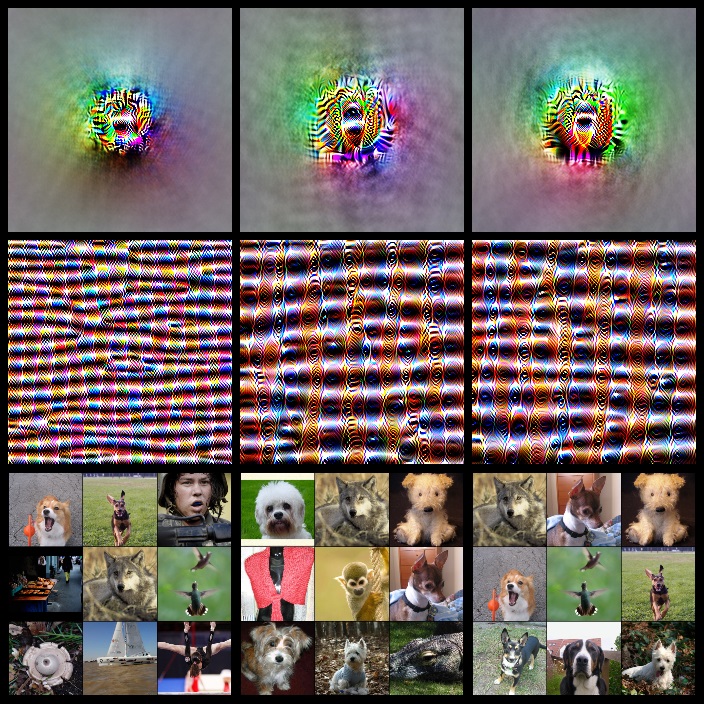}
        \textbf{2.1}: channel 13
        \vspace{0.15\linewidth}
    \end{subfigure}
    \begin{subfigure}{.16\linewidth}
        \centering
        \includegraphics[width=\linewidth]{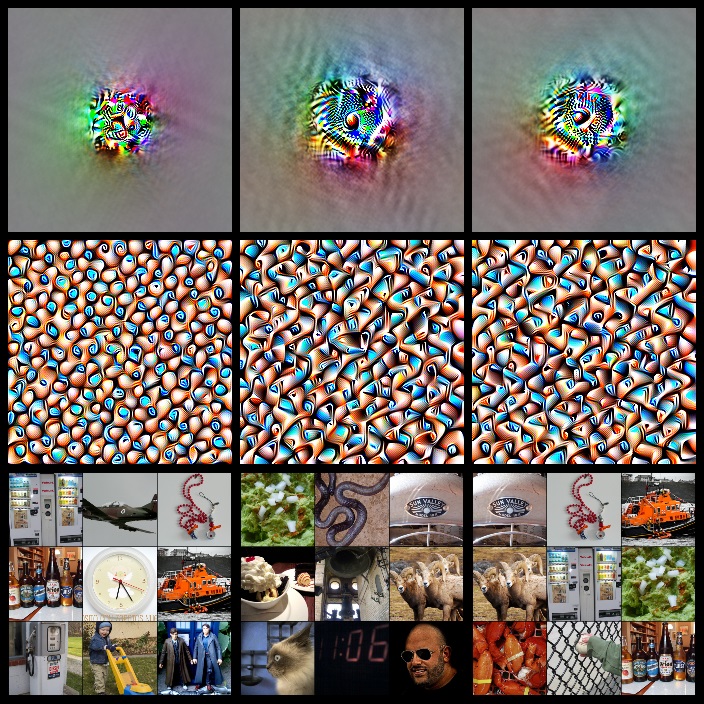}
        18
        \vspace{0.15\linewidth}
    \end{subfigure}
    \begin{subfigure}{.16\linewidth}
        \centering
        \includegraphics[width=\linewidth]{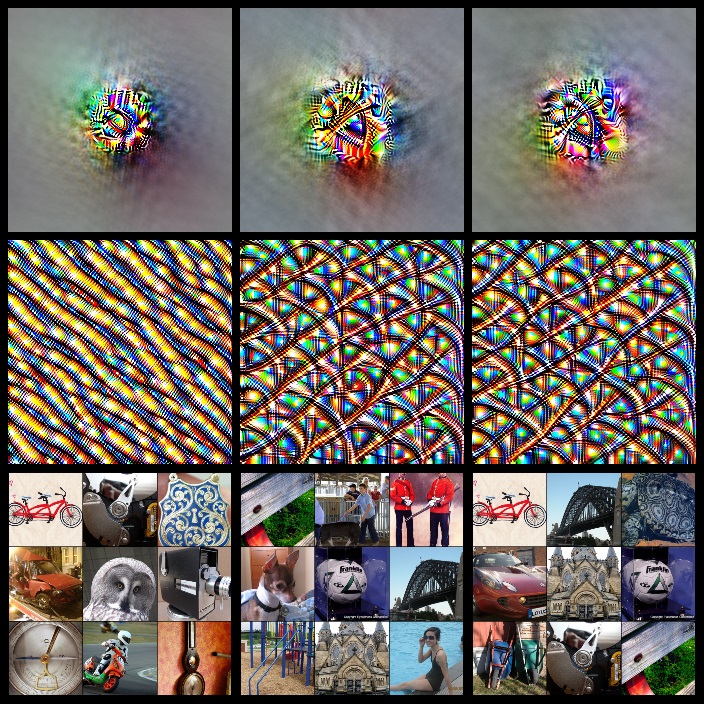}
        25
        \vspace{0.15\linewidth}
    \end{subfigure}
    \begin{subfigure}{.16\linewidth}
        \centering
        \includegraphics[width=\linewidth]{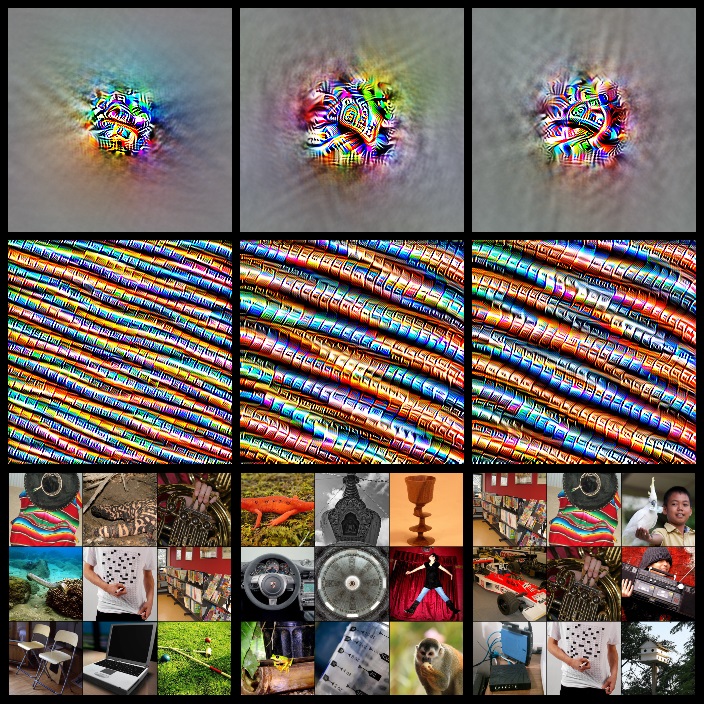}
        30
        \vspace{0.15\linewidth}
    \end{subfigure}
    \begin{subfigure}{.16\linewidth}
        \centering
        \includegraphics[width=\linewidth]{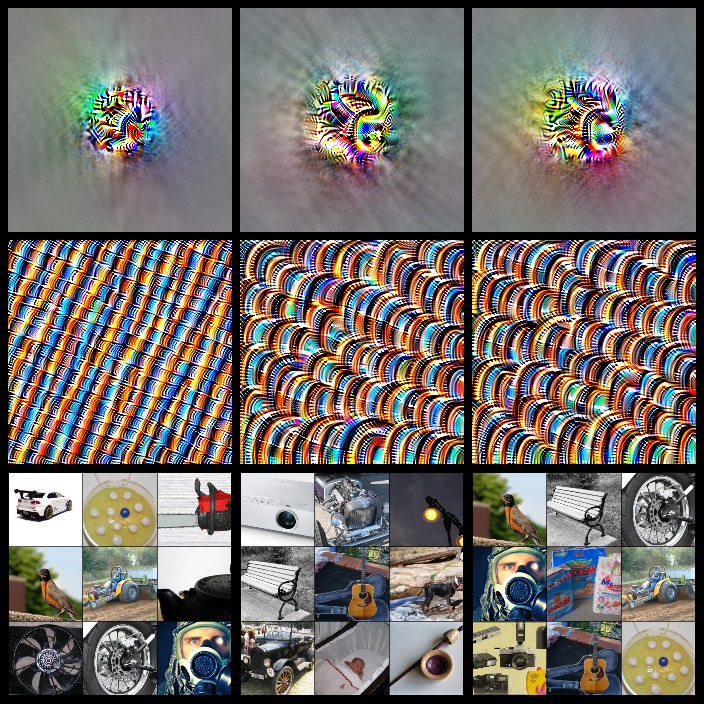}
        31
        \vspace{0.15\linewidth}
    \end{subfigure}
    \begin{subfigure}{.16\linewidth}
        \centering
        \includegraphics[width=\linewidth]{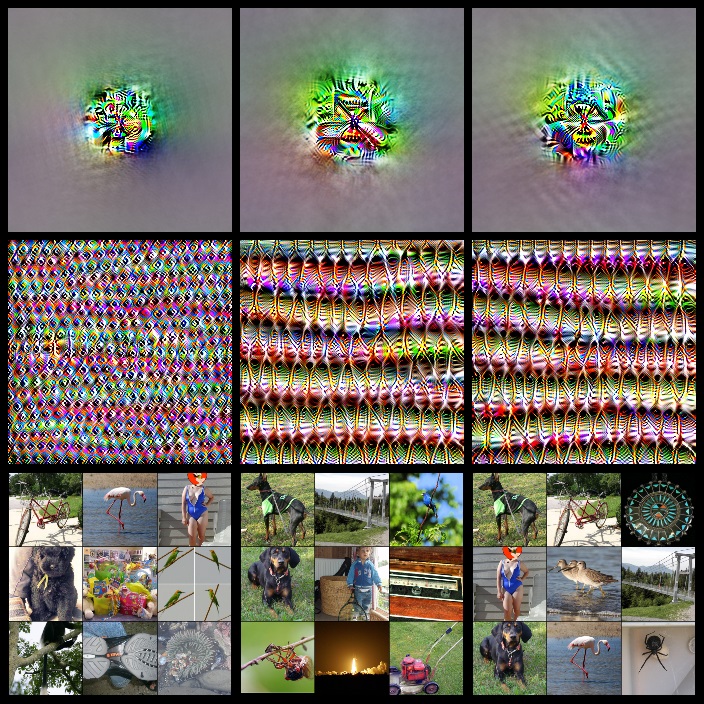}
        35
        \vspace{0.15\linewidth}
    \end{subfigure}

    \begin{subfigure}{.16\linewidth}
        \centering
        \includegraphics[width=\linewidth]{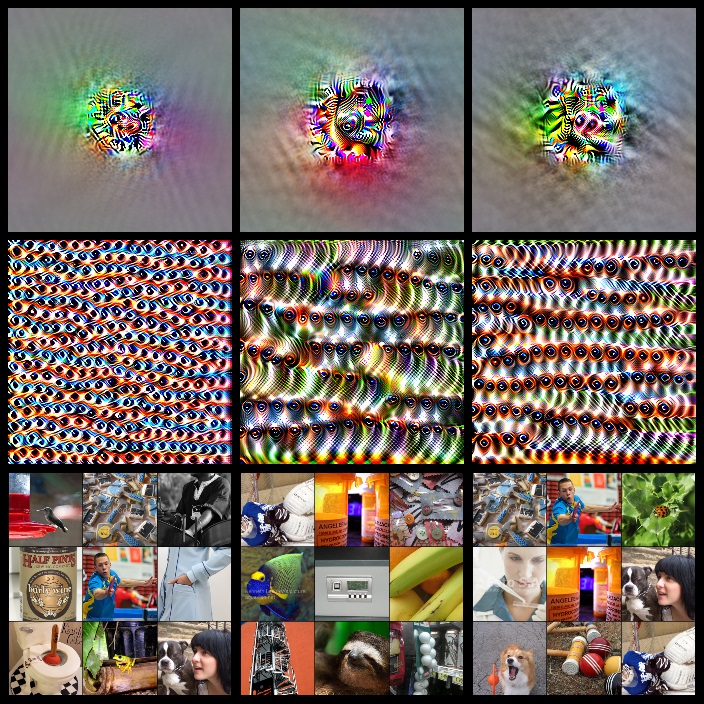}
        36
        \vspace{0.15\linewidth}
    \end{subfigure}
    \begin{subfigure}{.16\linewidth}
        \centering
        \includegraphics[width=\linewidth]{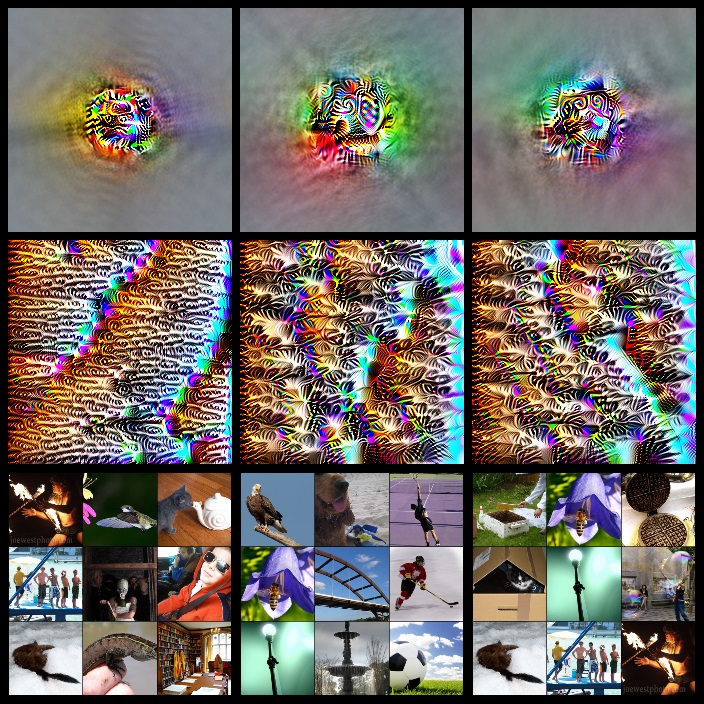}
        47
        \vspace{0.15\linewidth}
    \end{subfigure}
    \begin{subfigure}{.16\linewidth}
        \centering
        \includegraphics[width=\linewidth]{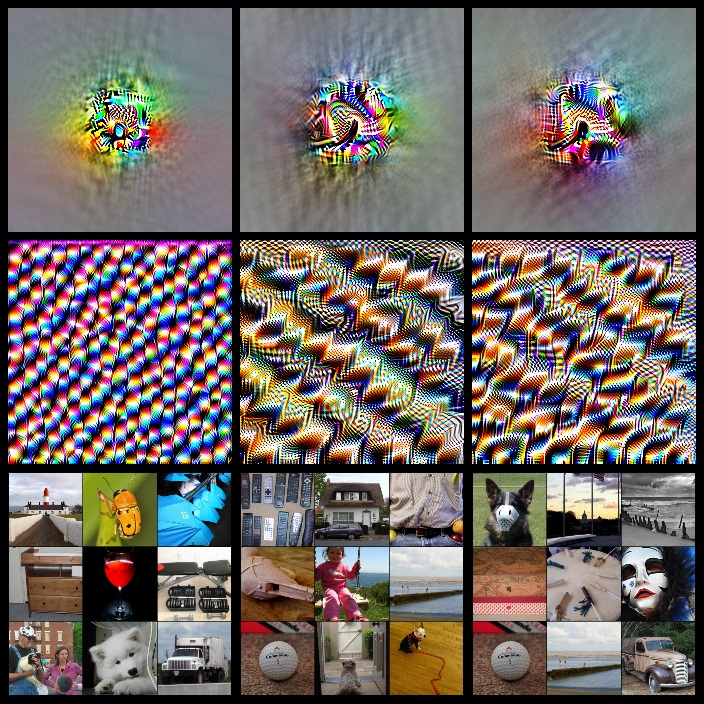}
        52
        \vspace{0.15\linewidth}
    \end{subfigure}
    \begin{subfigure}{.16\linewidth}
        \centering
        \includegraphics[width=\linewidth]{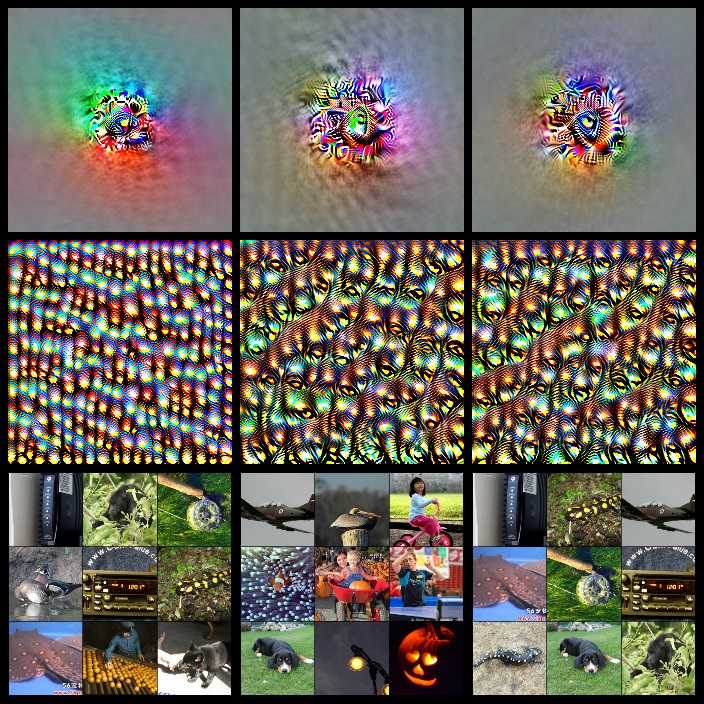}
        56
        \vspace{0.15\linewidth}
    \end{subfigure}
    \begin{subfigure}{.16\linewidth}
        \centering
        \includegraphics[width=\linewidth]{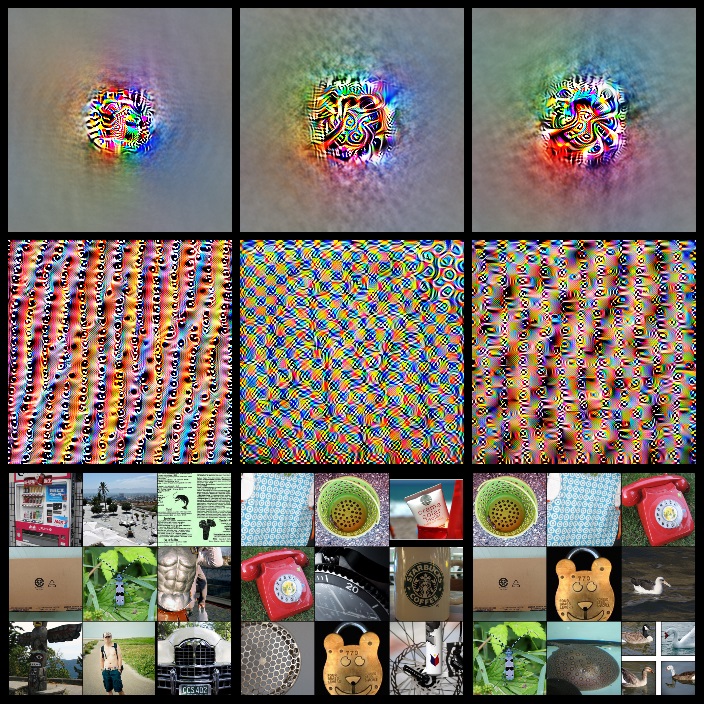}
        67
        \vspace{0.15\linewidth}
    \end{subfigure}
    \begin{subfigure}{.16\linewidth}
        \centering
        \includegraphics[width=\linewidth]{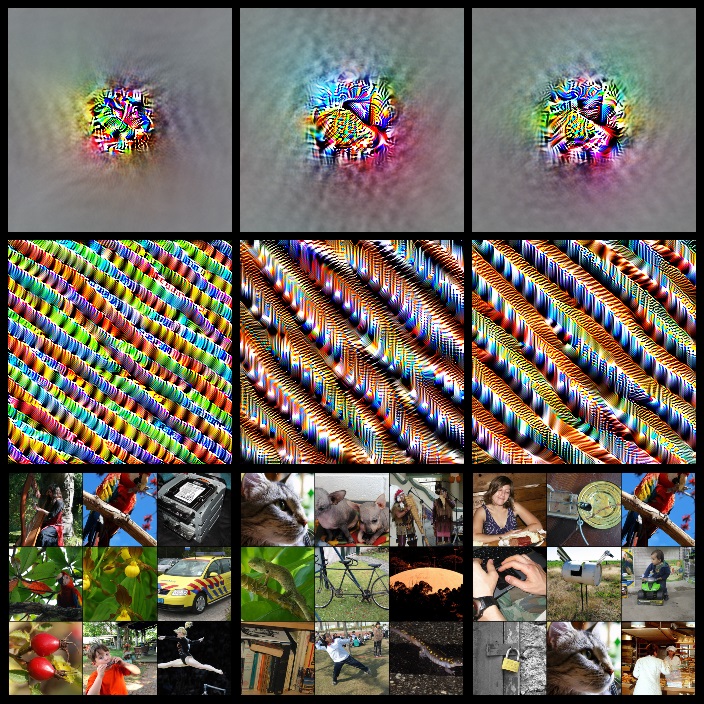}
        72
        \vspace{0.15\linewidth}
    \end{subfigure}

    \begin{subfigure}{.16\linewidth}
        \centering
        \includegraphics[width=\linewidth]{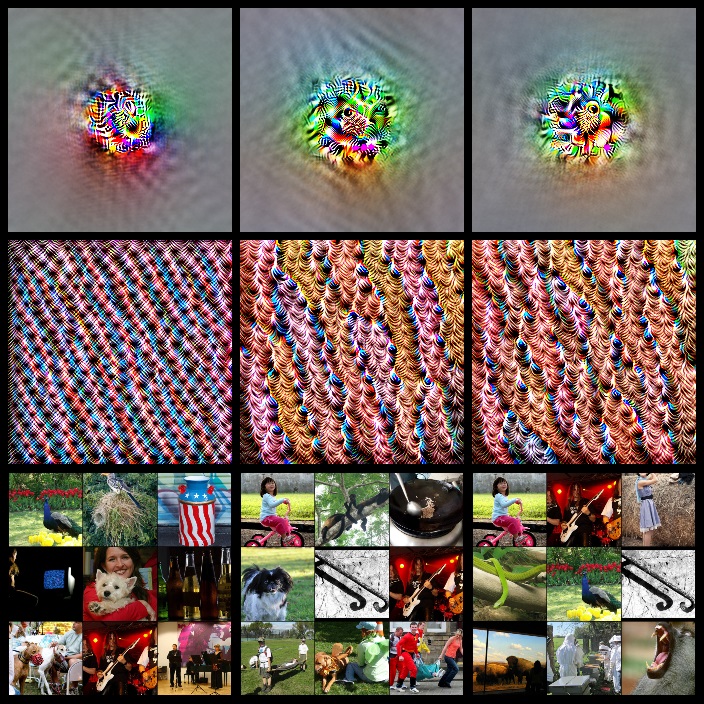}
        105
        \vspace{0.15\linewidth}
    \end{subfigure}
    \begin{subfigure}{.16\linewidth}
        \centering
        \includegraphics[width=\linewidth]{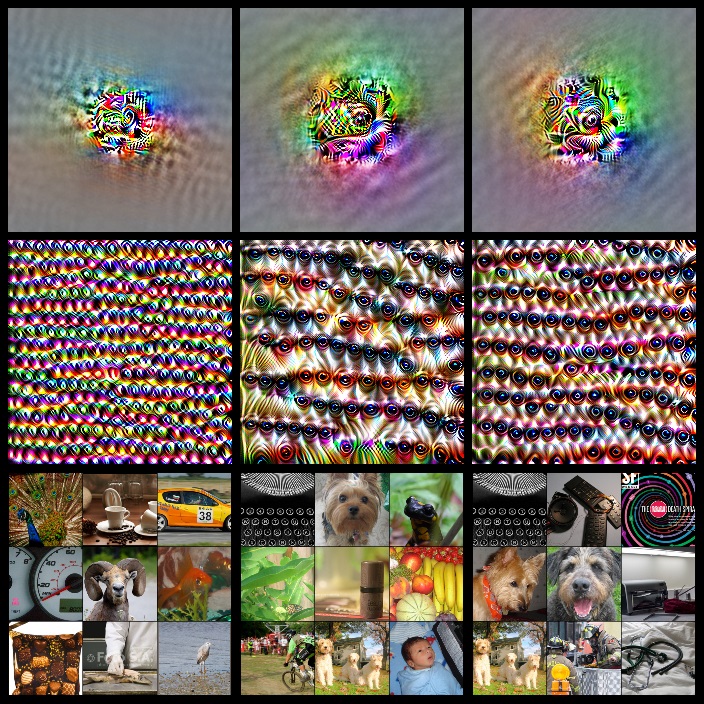}
        106
        \vspace{0.15\linewidth}
    \end{subfigure}
    \begin{subfigure}{.16\linewidth}
        \centering
        \includegraphics[width=\linewidth]{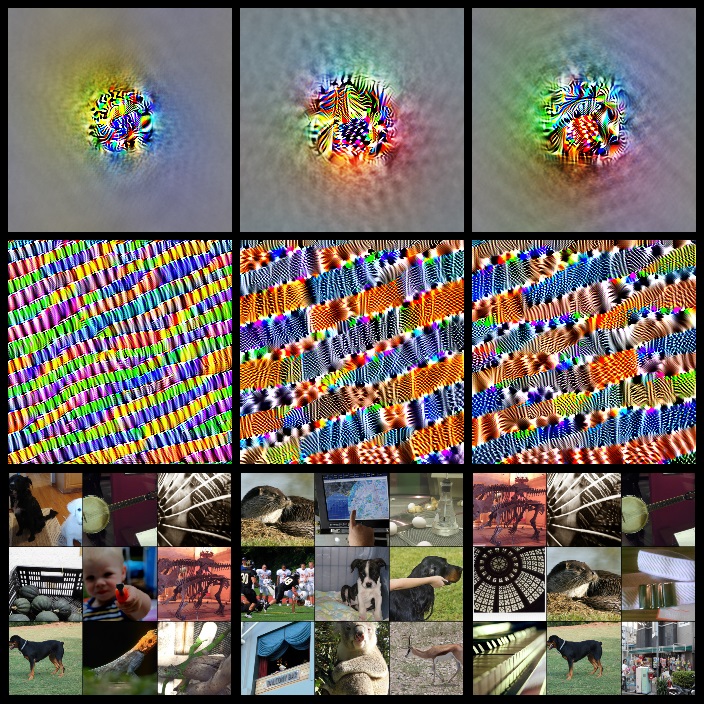}
        110
        \vspace{0.15\linewidth}
    \end{subfigure}
    \begin{subfigure}{.16\linewidth}
        \centering
        \includegraphics[width=\linewidth]{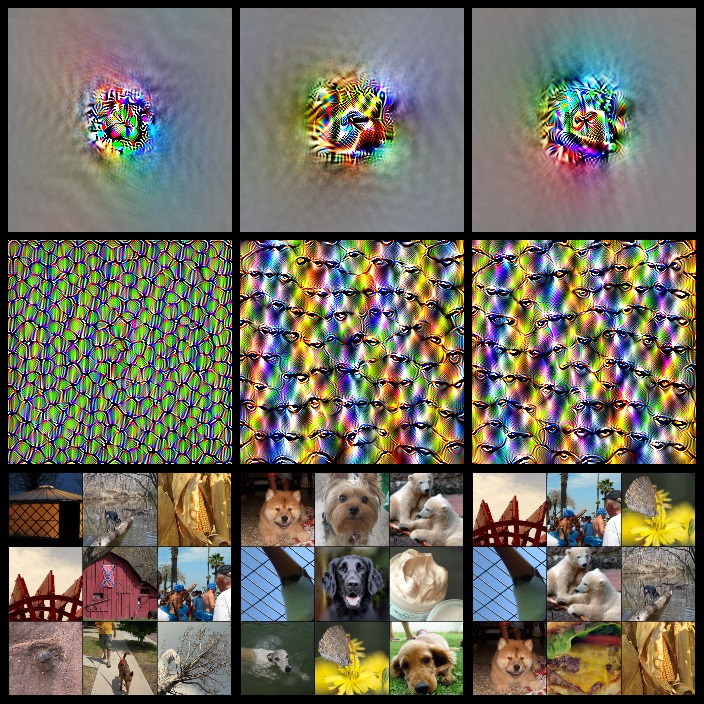}
        124
        \vspace{0.15\linewidth}
    \end{subfigure}
    \begin{subfigure}{.16\linewidth}
        \centering
        \includegraphics[width=\linewidth]{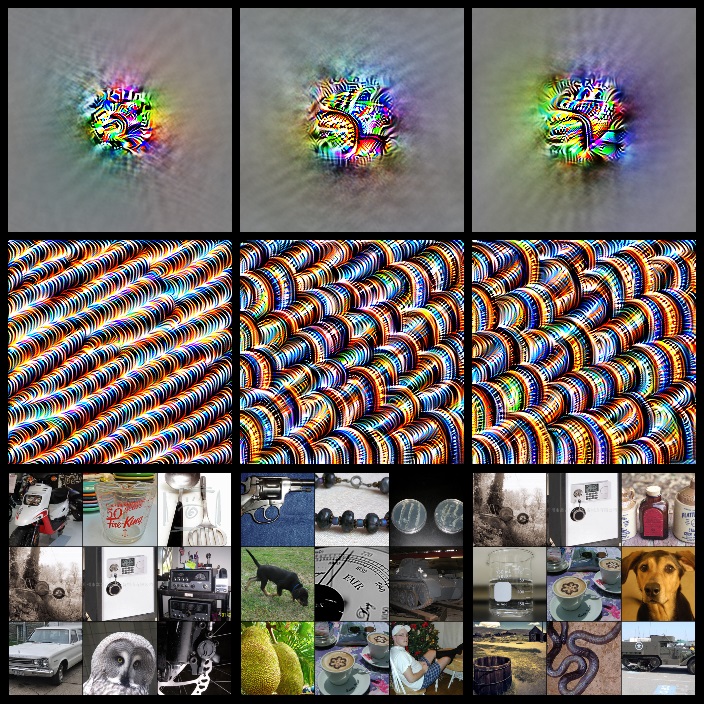}
        127
        \vspace{0.15\linewidth}
    \end{subfigure}
    
    \caption{Grids of maximally exciting images for all remaining scale invariant criteria-passing channels in blocks 1.1, 2.0, and 2.1.  All channels are zero-indexed.}
    \label{fig:A1}
\end{figure*}

\begin{figure*}[t]
    \centering
    \begin{subfigure}{.12\linewidth}
        \centering
        \includegraphics[width=\linewidth]{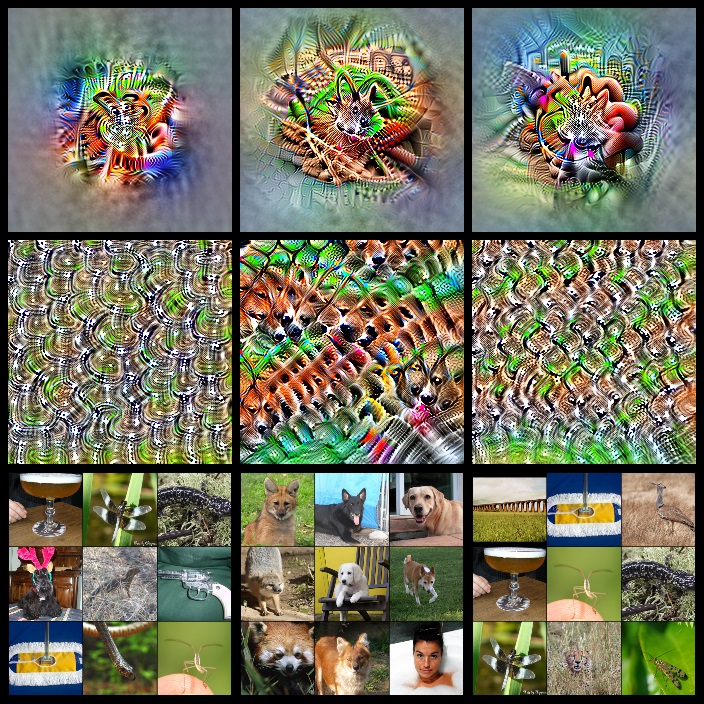}
        \textbf{3.1}: channel 1
        \vspace{0.15\linewidth}
    \end{subfigure}
    \begin{subfigure}{.12\linewidth}
        \centering
        \includegraphics[width=\linewidth]{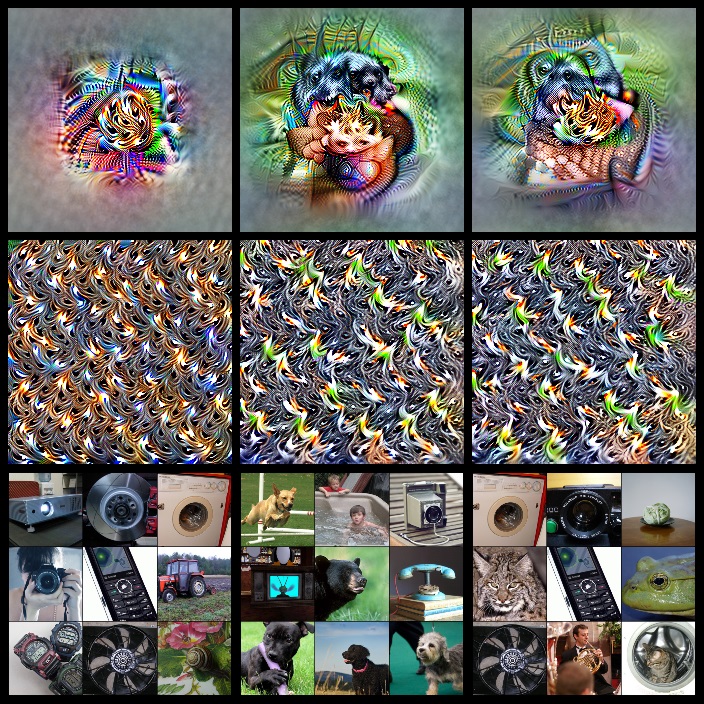}
        9
        \vspace{0.15\linewidth}
    \end{subfigure}
    \begin{subfigure}{.12\linewidth}
        \centering
        \includegraphics[width=\linewidth]{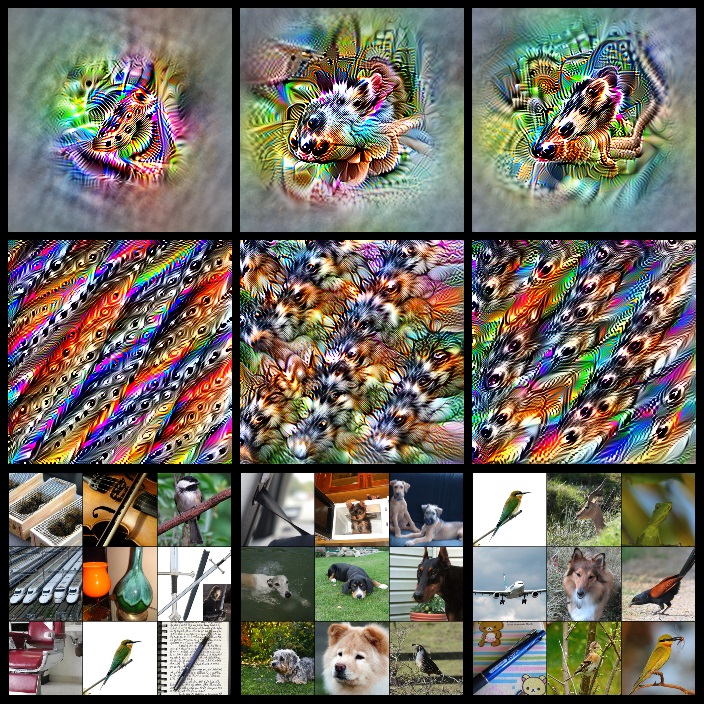}
        13
        \vspace{0.15\linewidth}
    \end{subfigure}
    \begin{subfigure}{.12\linewidth}
        \centering
        \includegraphics[width=\linewidth]{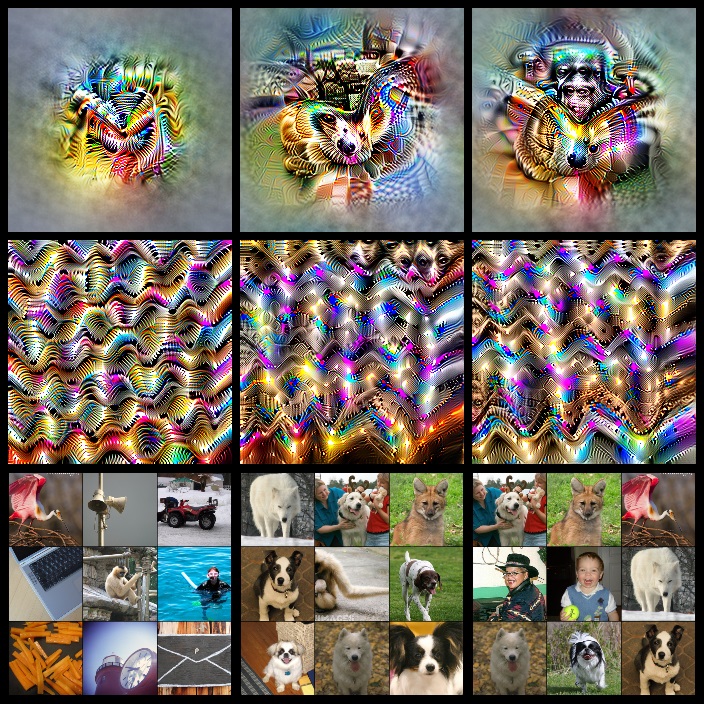}
        14
        \vspace{0.15\linewidth}
    \end{subfigure}
    \begin{subfigure}{.12\linewidth}
        \centering
        \includegraphics[width=\linewidth]{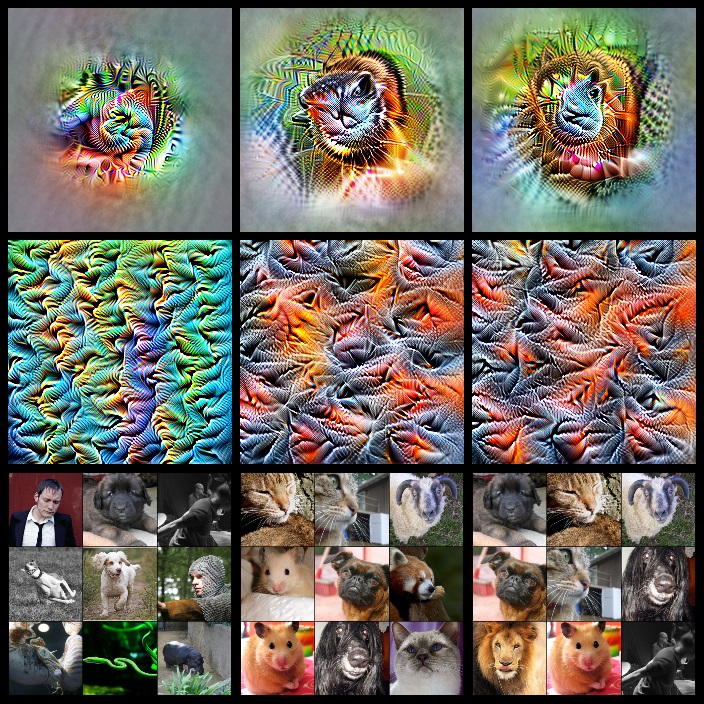}
        16
        \vspace{0.15\linewidth}
    \end{subfigure}
    \begin{subfigure}{.12\linewidth}
        \centering
        \includegraphics[width=\linewidth]{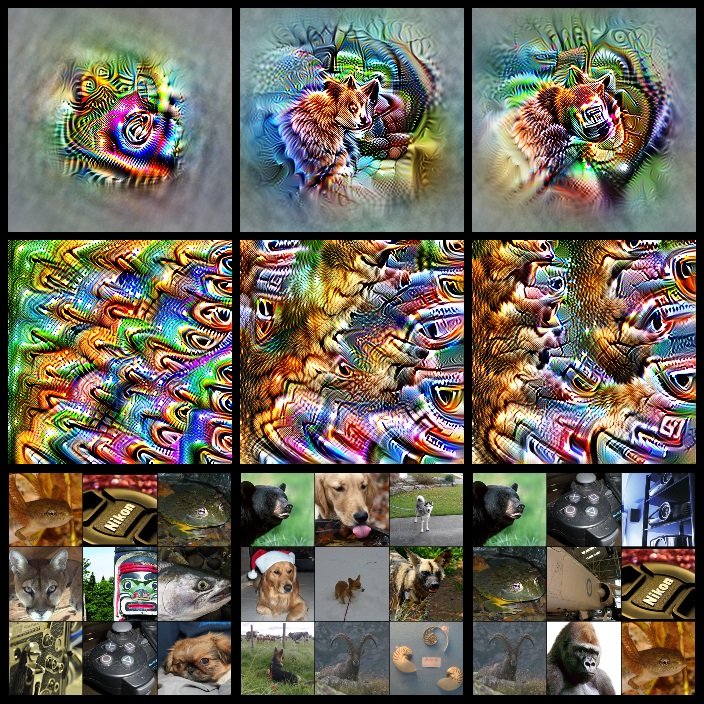}
        48
        \vspace{0.15\linewidth}
    \end{subfigure}
    \begin{subfigure}{.12\linewidth}
        \centering
        \includegraphics[width=\linewidth]{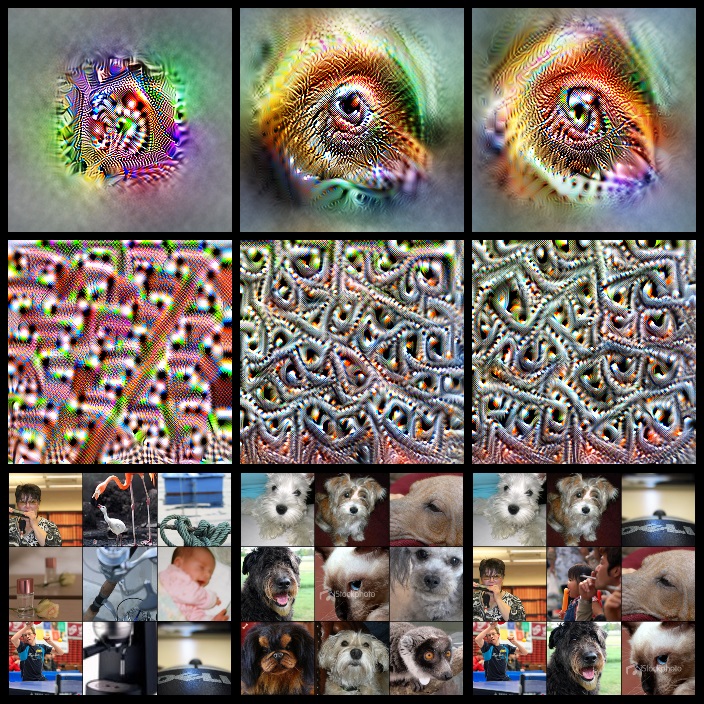}
        58
        \vspace{0.15\linewidth}
    \end{subfigure}
    \begin{subfigure}{.12\linewidth}
        \centering
        \includegraphics[width=\linewidth]{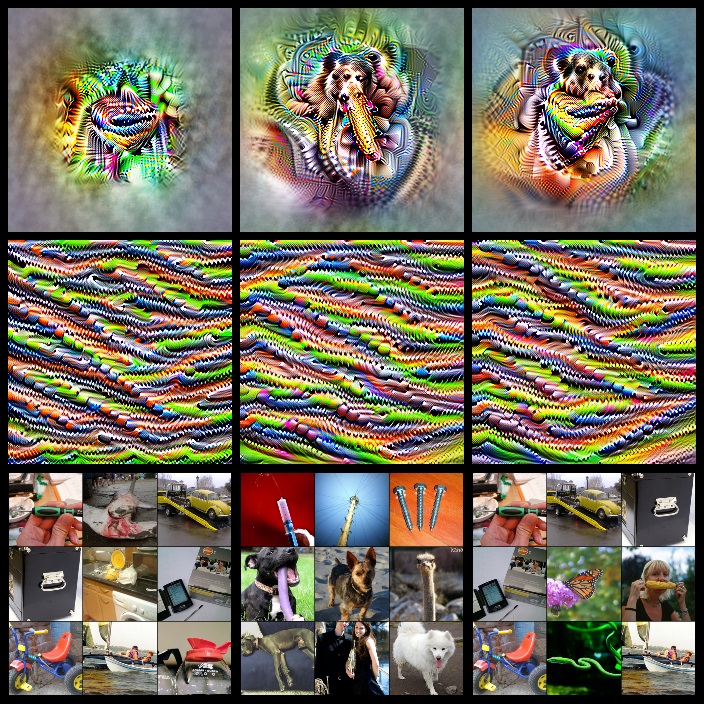}
        59
        \vspace{0.15\linewidth}
    \end{subfigure}
    
    \begin{subfigure}{.12\linewidth}
        \centering
        \includegraphics[width=\linewidth]{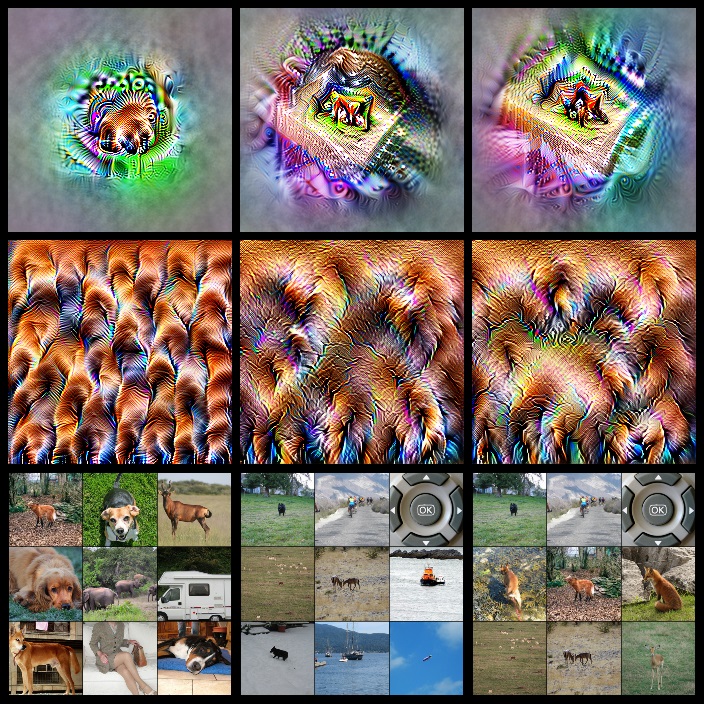}
        63
        \vspace{0.15\linewidth}
    \end{subfigure}
    \begin{subfigure}{.12\linewidth}
        \centering
        \includegraphics[width=\linewidth]{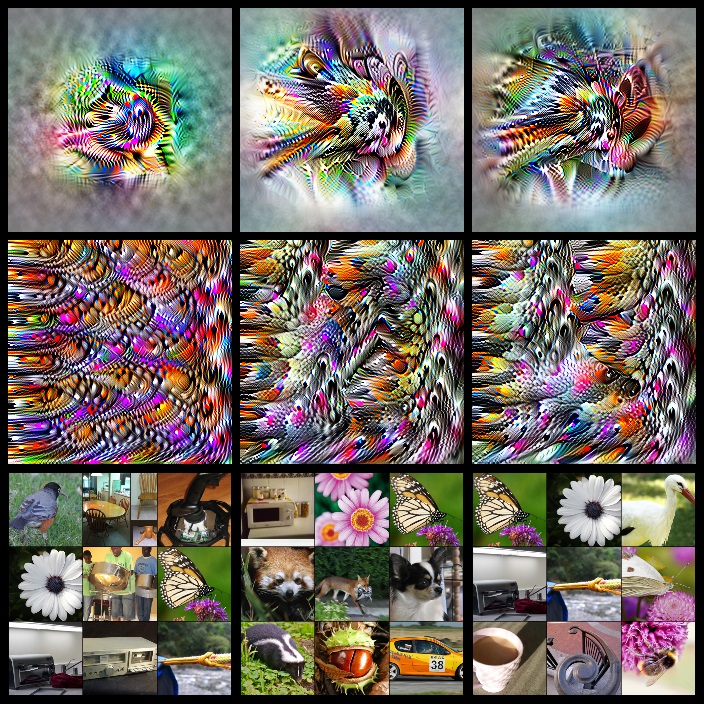}
        67
        \vspace{0.15\linewidth}
    \end{subfigure}
    \begin{subfigure}{.12\linewidth}
        \centering
        \includegraphics[width=\linewidth]{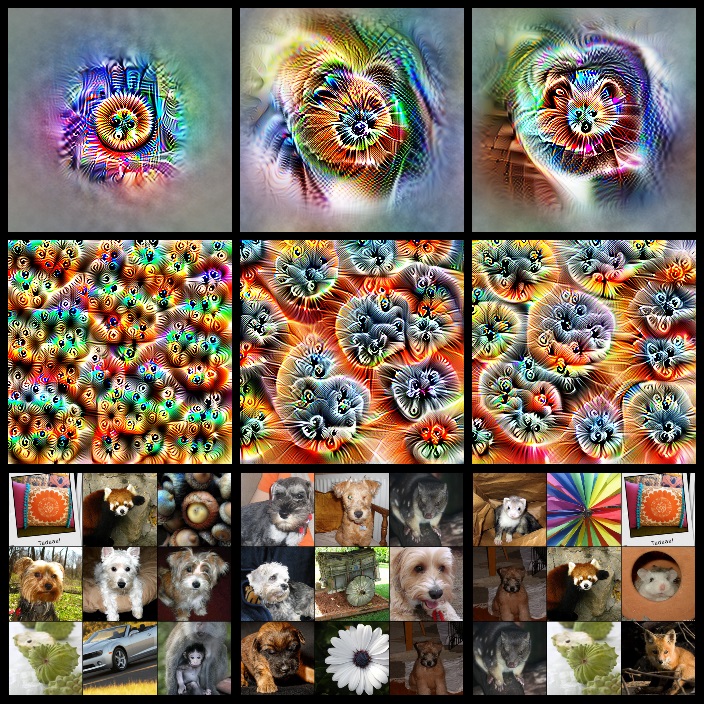}
        69
        \vspace{0.15\linewidth}
    \end{subfigure}
    \begin{subfigure}{.12\linewidth}
        \centering
        \includegraphics[width=\linewidth]{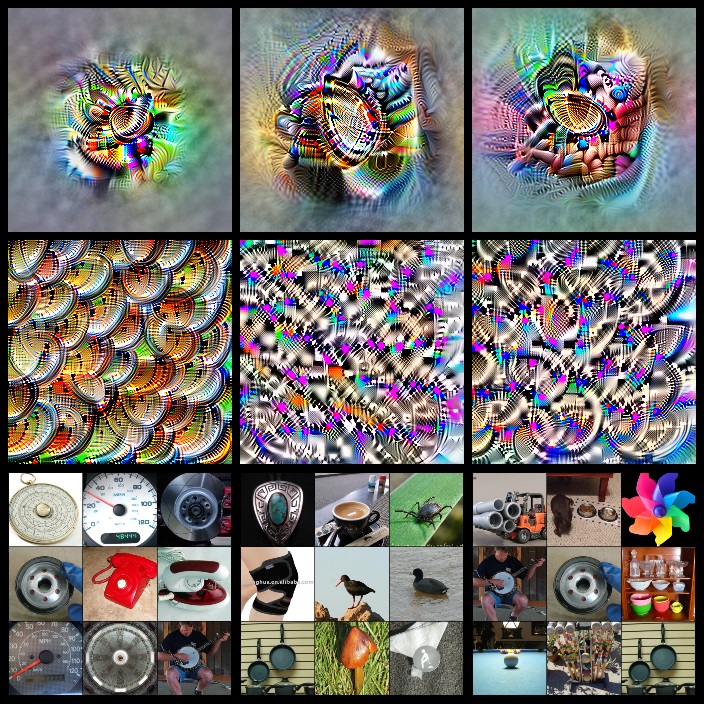}
        88
        \vspace{0.15\linewidth}
    \end{subfigure}
    \begin{subfigure}{.12\linewidth}
        \centering
        \includegraphics[width=\linewidth]{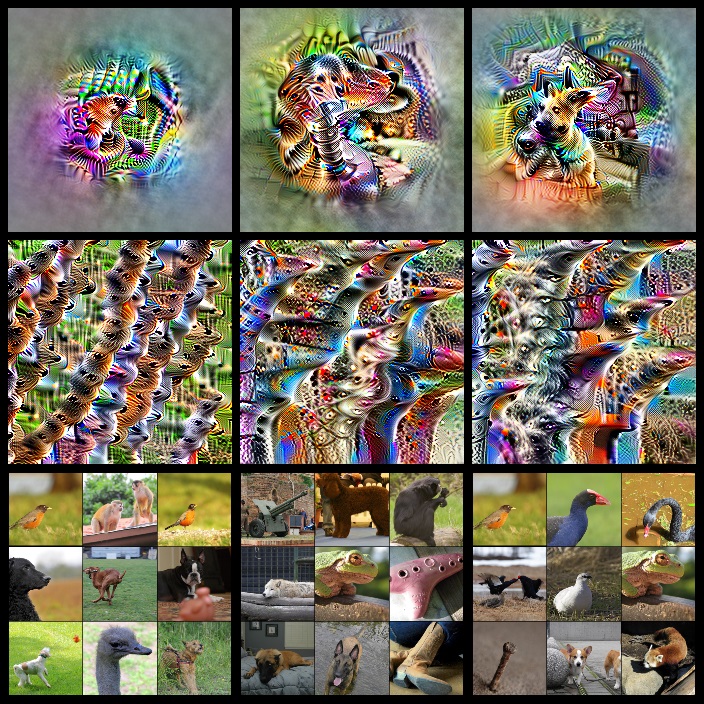}
        100
        \vspace{0.15\linewidth}
    \end{subfigure}
    \begin{subfigure}{.12\linewidth}
        \centering
        \includegraphics[width=\linewidth]{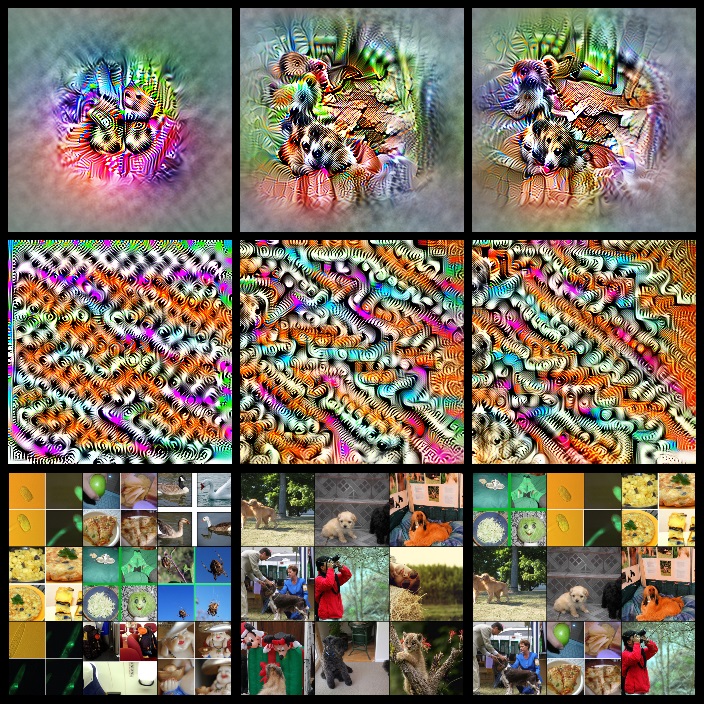}
        117
        \vspace{0.15\linewidth}
    \end{subfigure}
    \begin{subfigure}{.12\linewidth}
        \centering
        \includegraphics[width=\linewidth]{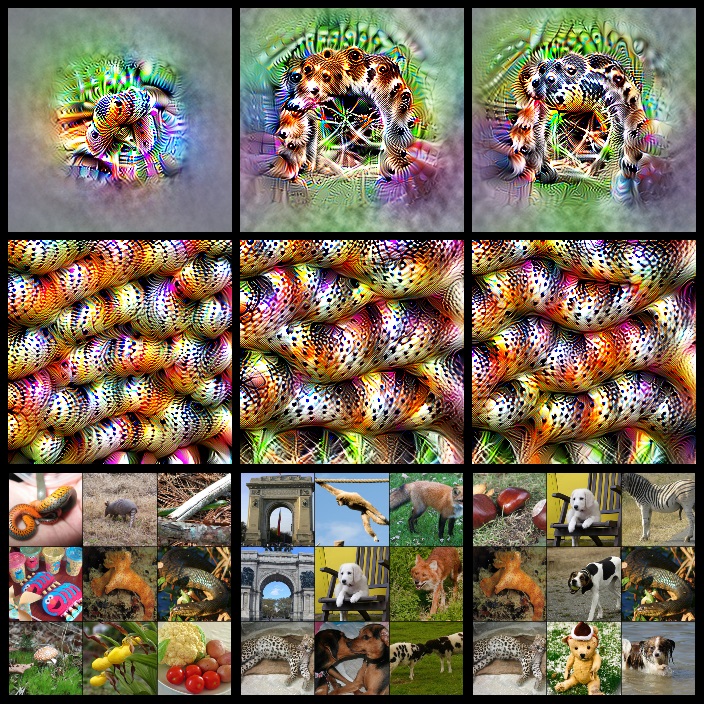}
        126
        \vspace{0.15\linewidth}
    \end{subfigure}
    \begin{subfigure}{.12\linewidth}
        \centering
        \includegraphics[width=\linewidth]{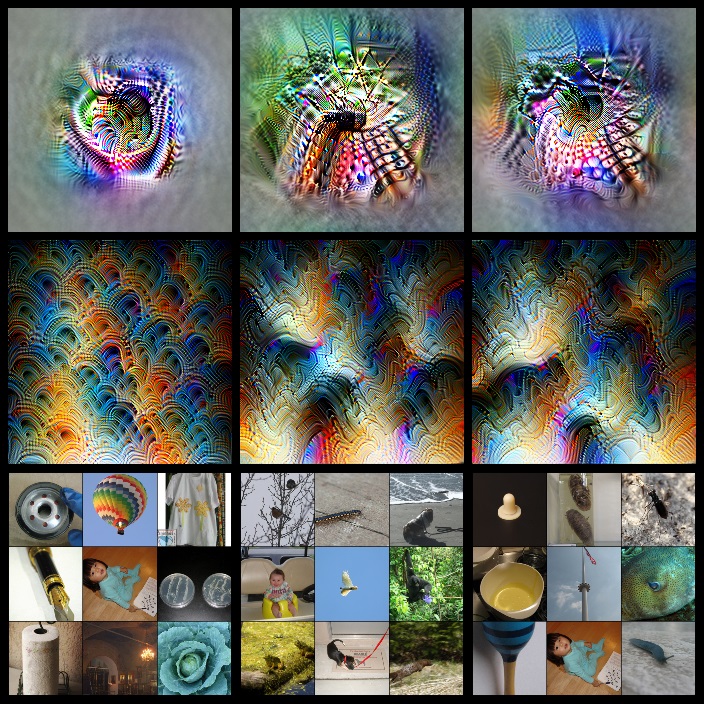}
        132
        \vspace{0.15\linewidth}
    \end{subfigure}
    
    \begin{subfigure}{.12\linewidth}
        \centering
        \includegraphics[width=\linewidth]{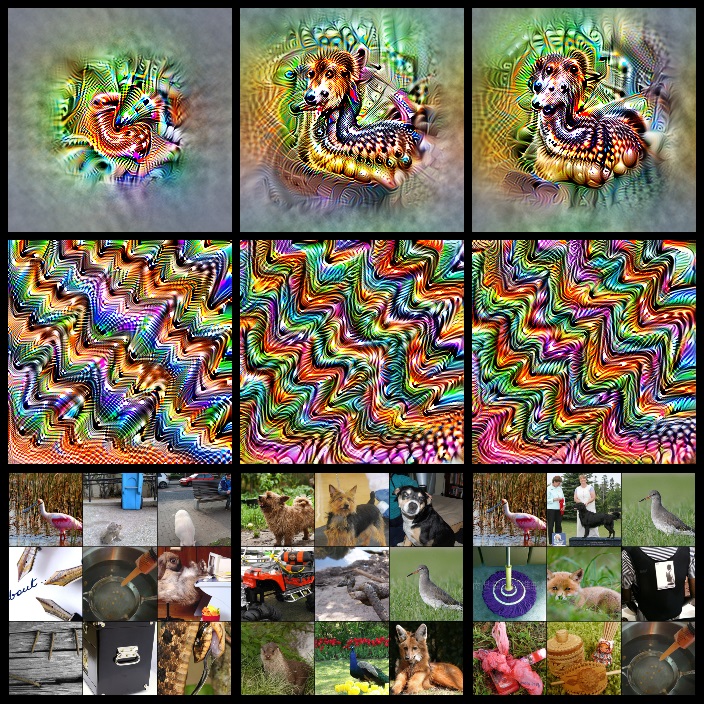}
        135
        \vspace{0.15\linewidth}
    \end{subfigure}
    \begin{subfigure}{.12\linewidth}
        \centering
        \includegraphics[width=\linewidth]{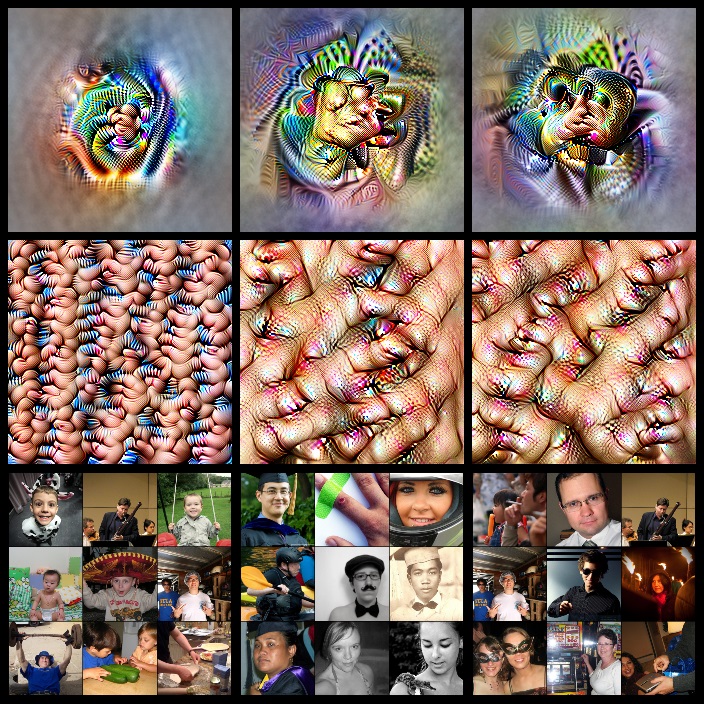}
        137
        \vspace{0.15\linewidth}
    \end{subfigure}
    \begin{subfigure}{.12\linewidth}
        \centering
        \includegraphics[width=\linewidth]{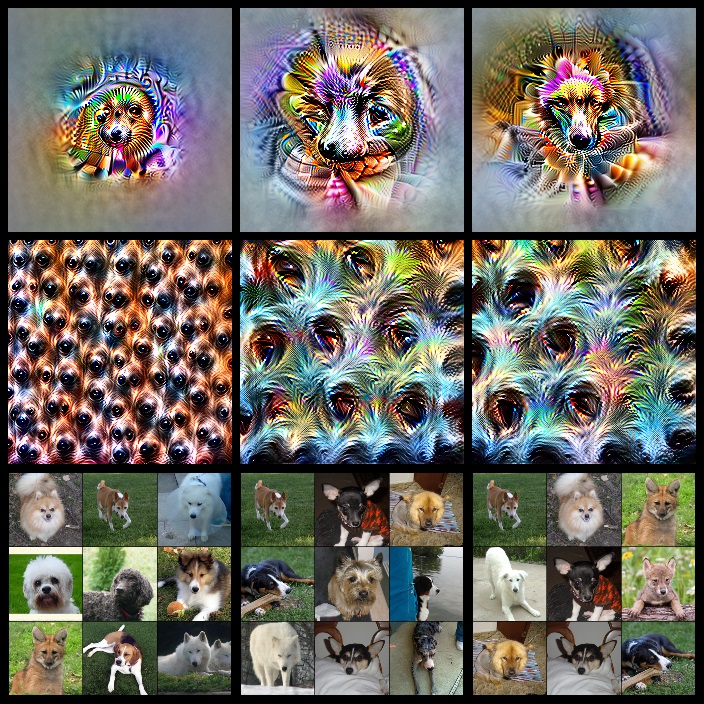}
        141
        \vspace{0.15\linewidth}
    \end{subfigure}
    \begin{subfigure}{.12\linewidth}
        \centering
        \includegraphics[width=\linewidth]{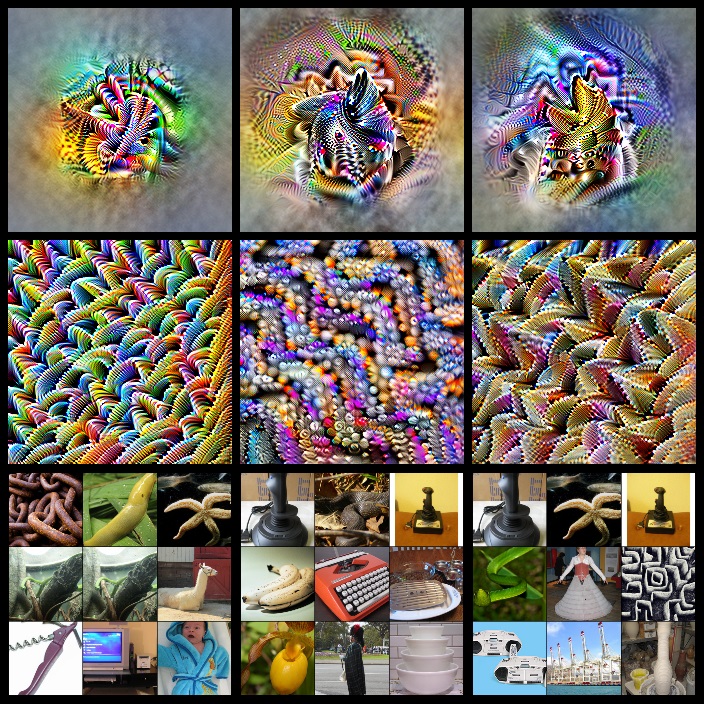}
        144
        \vspace{0.15\linewidth}
    \end{subfigure}
    \begin{subfigure}{.12\linewidth}
        \centering
        \includegraphics[width=\linewidth]{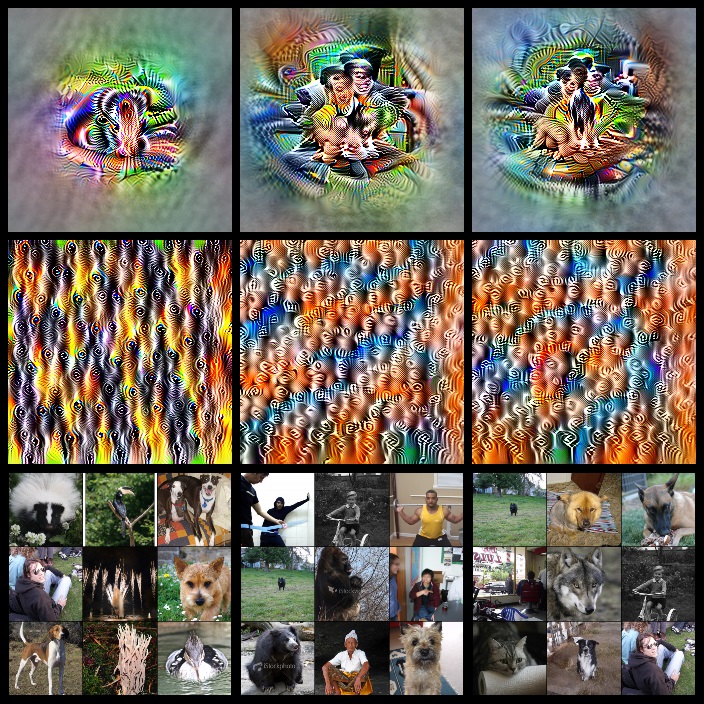}
        147
        \vspace{0.15\linewidth}
    \end{subfigure}
    \begin{subfigure}{.12\linewidth}
        \centering
        \includegraphics[width=\linewidth]{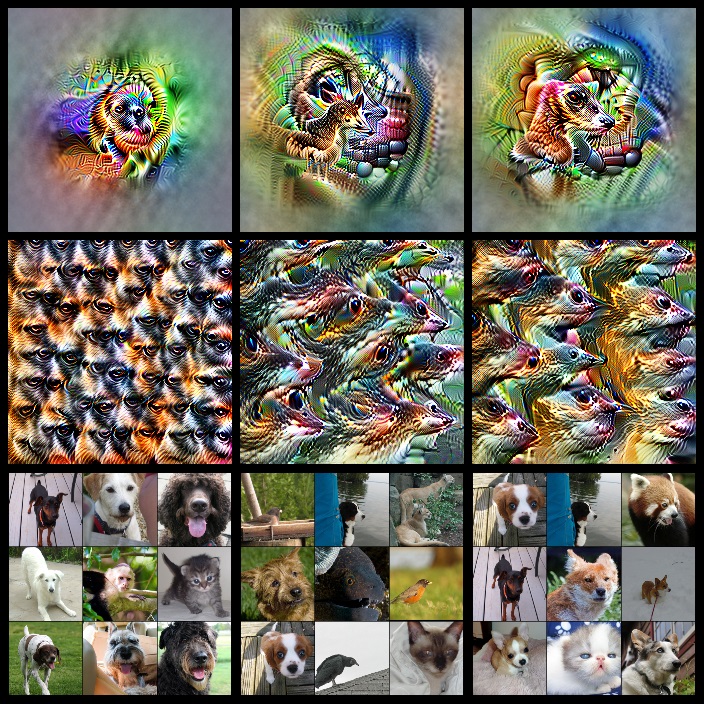}
        149
        \vspace{0.15\linewidth}
    \end{subfigure}
    \begin{subfigure}{.12\linewidth}
        \centering
        \includegraphics[width=\linewidth]{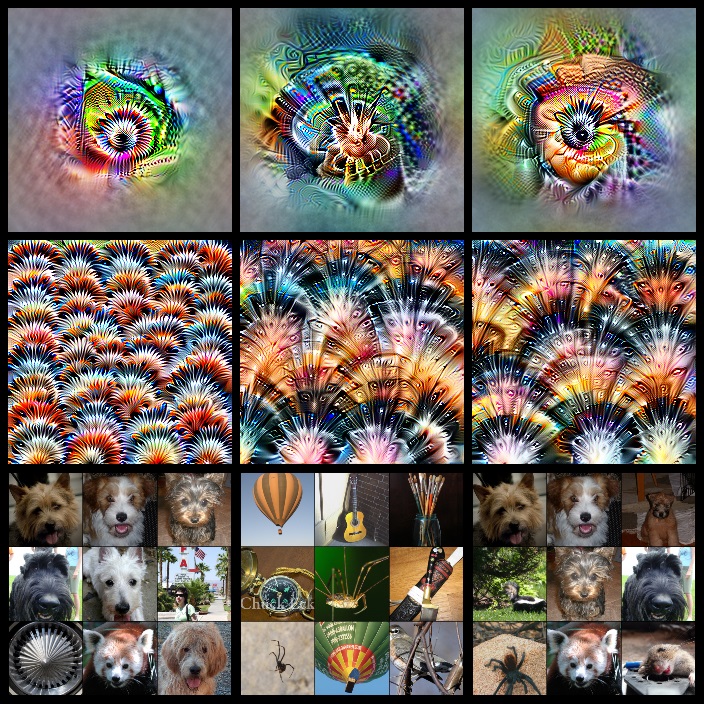}
        153
        \vspace{0.15\linewidth}
    \end{subfigure}
    \begin{subfigure}{.12\linewidth}
        \centering
        \includegraphics[width=\linewidth]{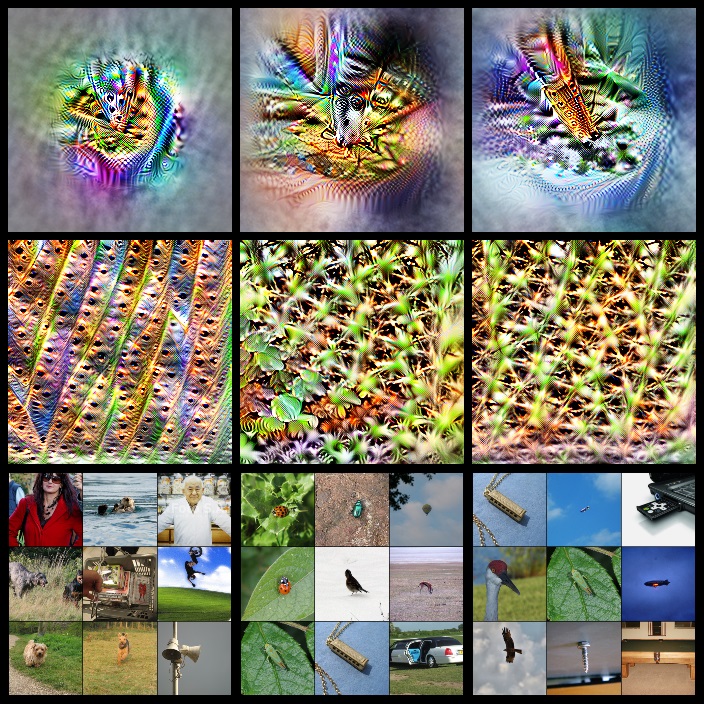}
        158
        \vspace{0.15\linewidth}
    \end{subfigure}
    
    \begin{subfigure}{.12\linewidth}
        \centering
        \includegraphics[width=\linewidth]{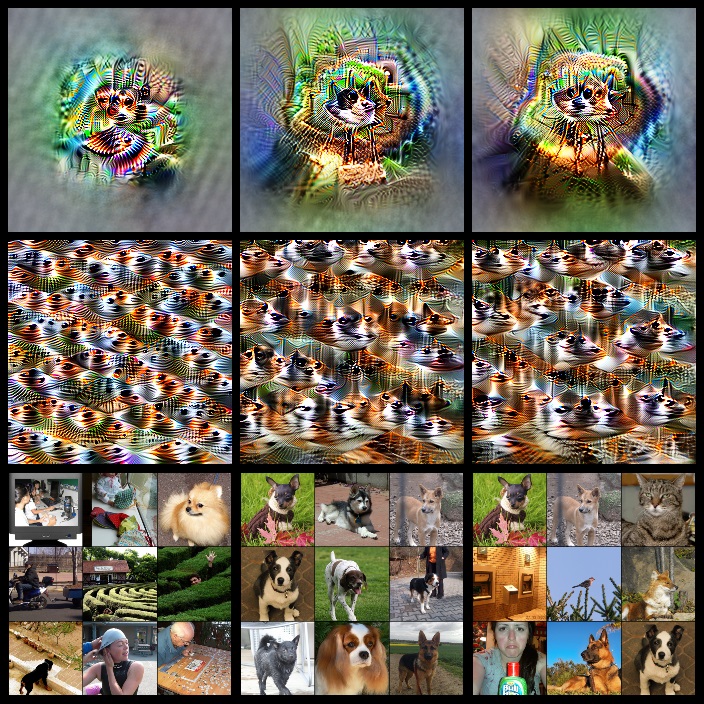}
        161
        \vspace{0.15\linewidth}
    \end{subfigure}
    \begin{subfigure}{.12\linewidth}
        \centering
        \includegraphics[width=\linewidth]{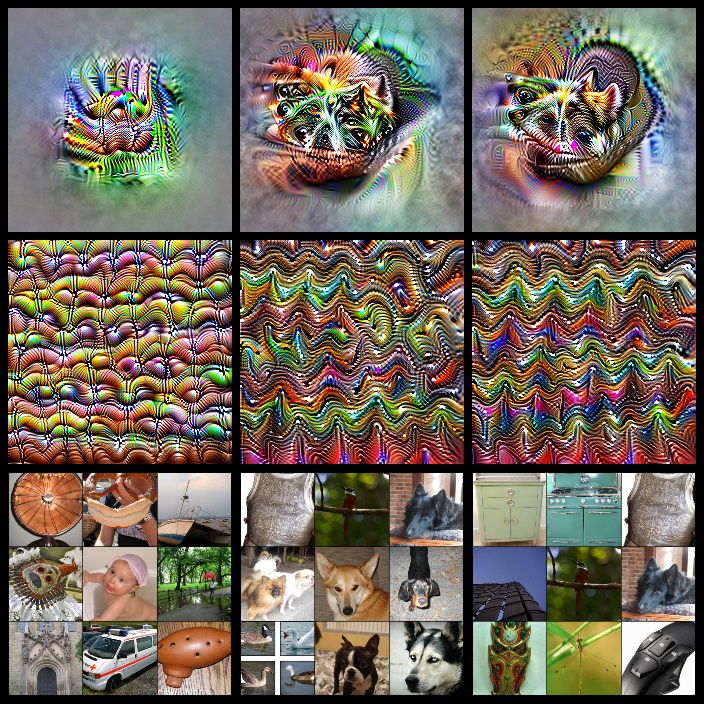}
        163
        \vspace{0.15\linewidth}
    \end{subfigure}
    \begin{subfigure}{.12\linewidth}
        \centering
        \includegraphics[width=\linewidth]{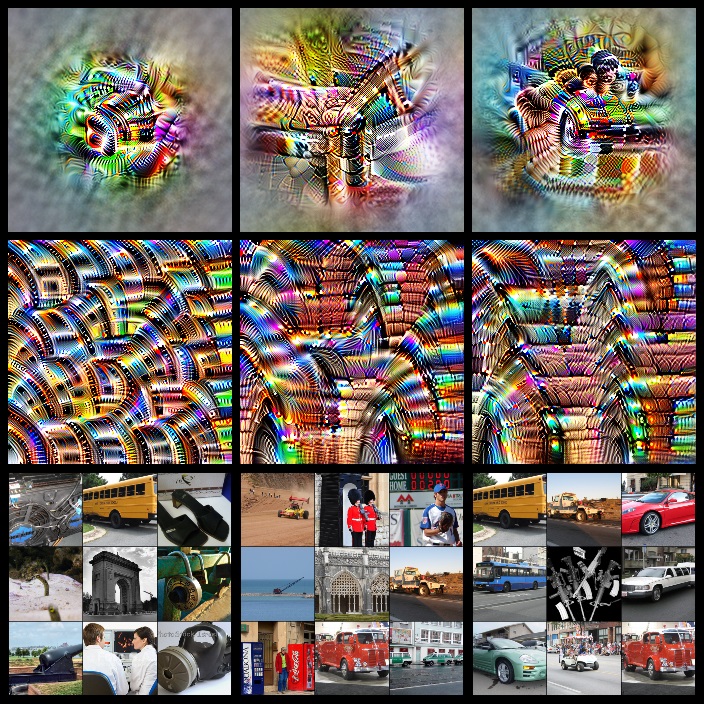}
        164
        \vspace{0.15\linewidth}
    \end{subfigure}
    \begin{subfigure}{.12\linewidth}
        \centering
        \includegraphics[width=\linewidth]{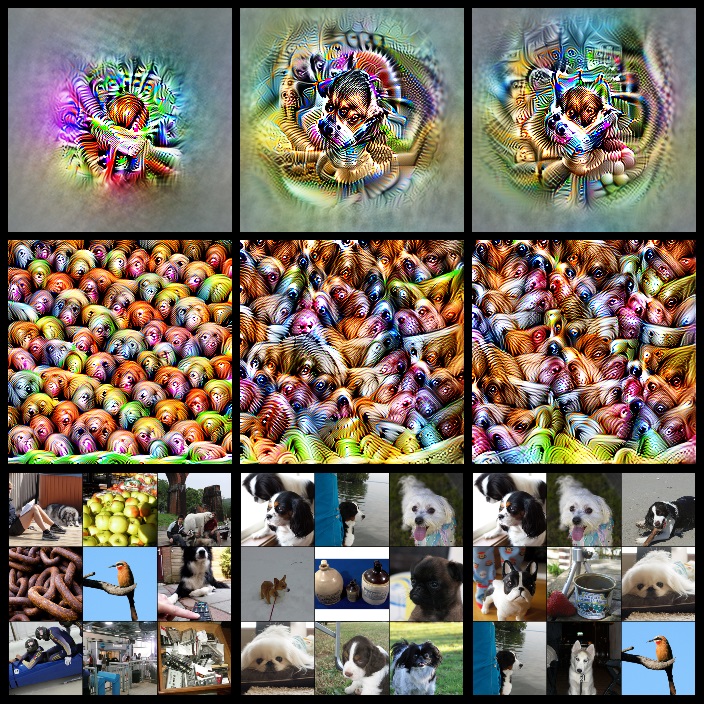}
        168
        \vspace{0.15\linewidth}
    \end{subfigure}
    \begin{subfigure}{.12\linewidth}
        \centering
        \includegraphics[width=\linewidth]{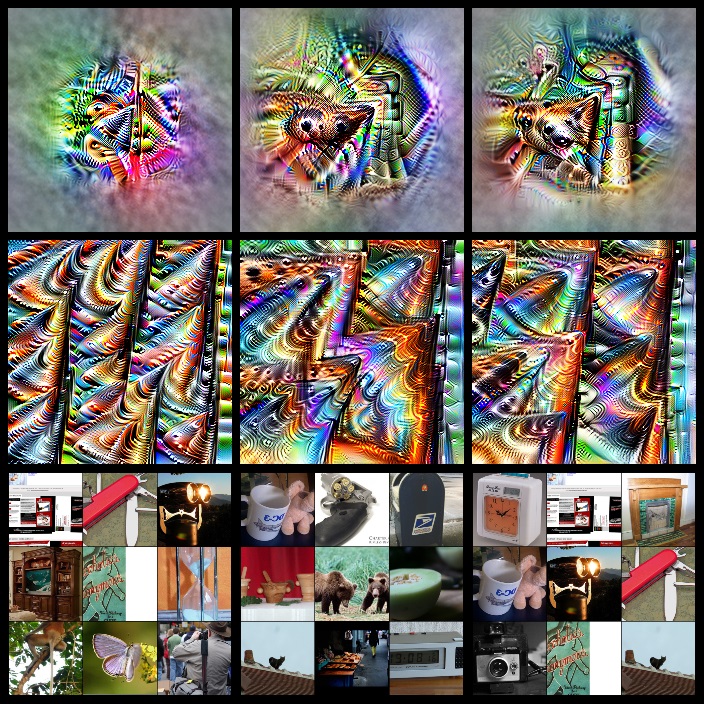}
        179
        \vspace{0.15\linewidth}
    \end{subfigure}
    \begin{subfigure}{.12\linewidth}
        \centering
        \includegraphics[width=\linewidth]{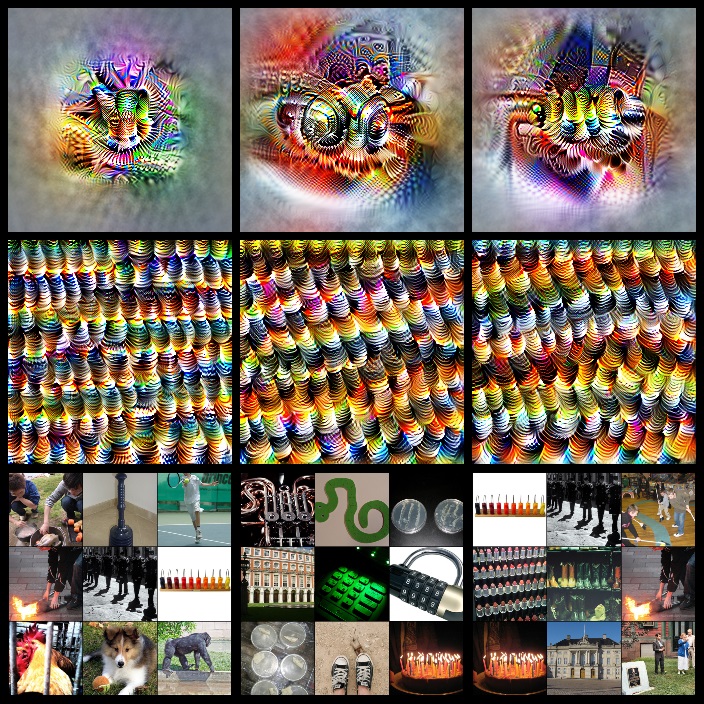}
        180
        \vspace{0.15\linewidth}
    \end{subfigure}
    \begin{subfigure}{.12\linewidth}
        \centering
        \includegraphics[width=\linewidth]{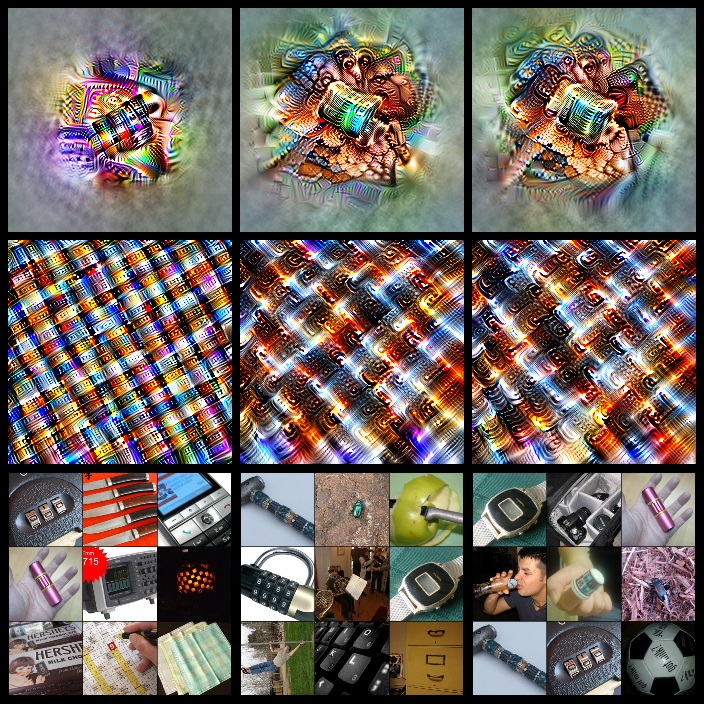}
        187
        \vspace{0.15\linewidth}
    \end{subfigure}
    \begin{subfigure}{.12\linewidth}
        \centering
        \includegraphics[width=\linewidth]{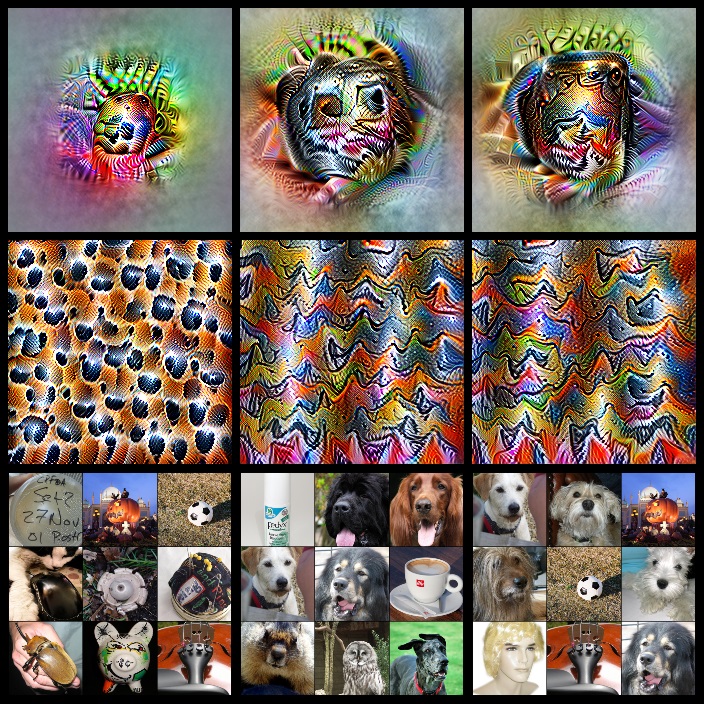}
        188
        \vspace{0.15\linewidth}
    \end{subfigure}
    
    \begin{subfigure}{.12\linewidth}
        \centering
        \includegraphics[width=\linewidth]{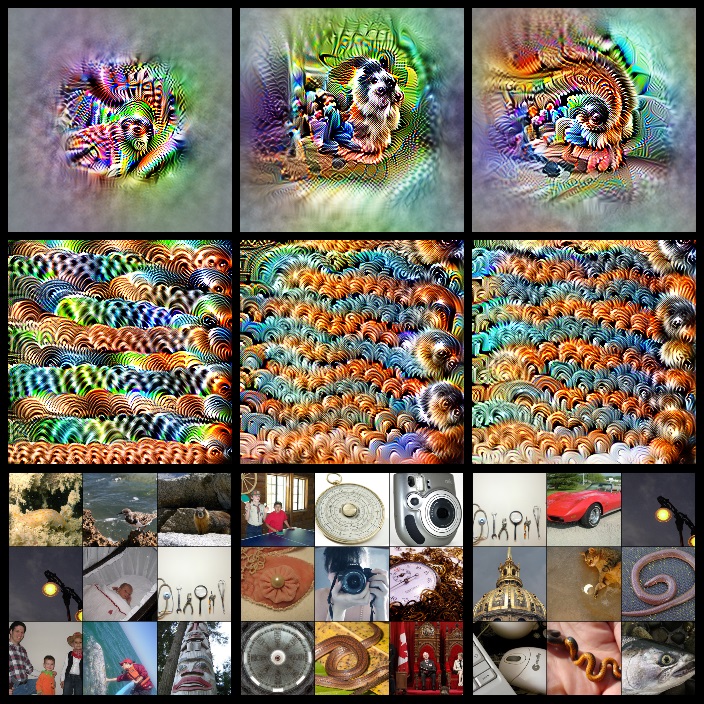}
        192
        \vspace{0.15\linewidth}
    \end{subfigure}
    \begin{subfigure}{.12\linewidth}
        \centering
        \includegraphics[width=\linewidth]{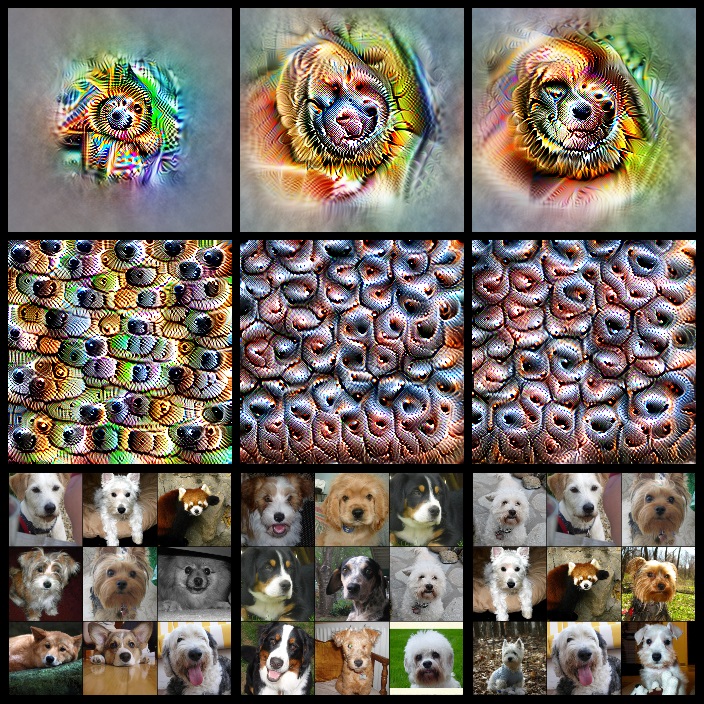}
        198
        \vspace{0.15\linewidth}
    \end{subfigure}
    \begin{subfigure}{.12\linewidth}
        \centering
        \includegraphics[width=\linewidth]{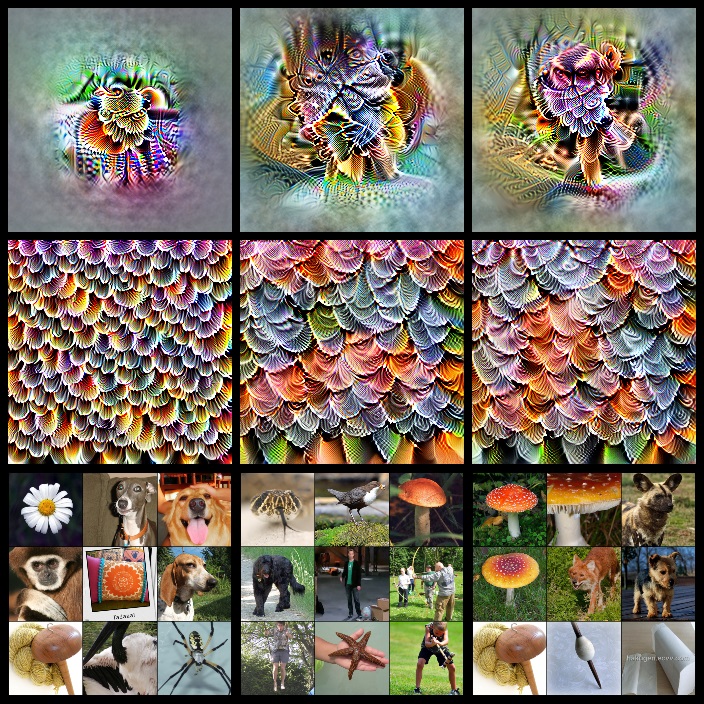}
        200
        \vspace{0.15\linewidth}
    \end{subfigure}
    \begin{subfigure}{.12\linewidth}
        \centering
        \includegraphics[width=\linewidth]{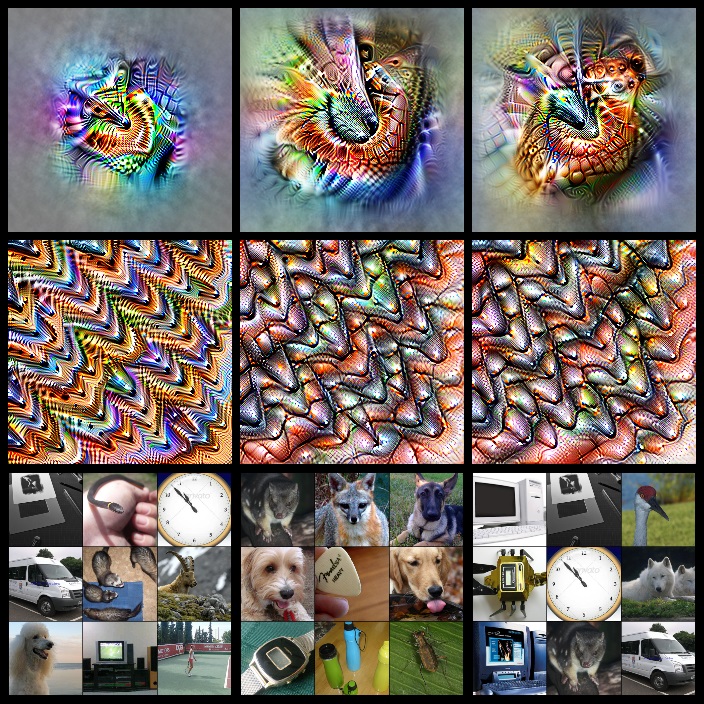}
        204
        \vspace{0.15\linewidth}
    \end{subfigure}
    \begin{subfigure}{.12\linewidth}
        \centering
        \includegraphics[width=\linewidth]{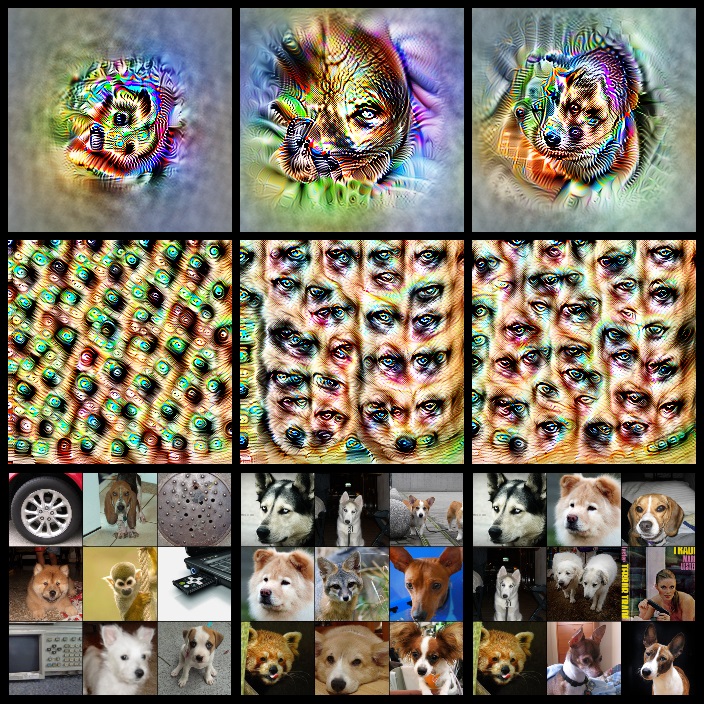}
        206
        \vspace{0.15\linewidth}
    \end{subfigure}
    \begin{subfigure}{.12\linewidth}
        \centering
        \includegraphics[width=\linewidth]{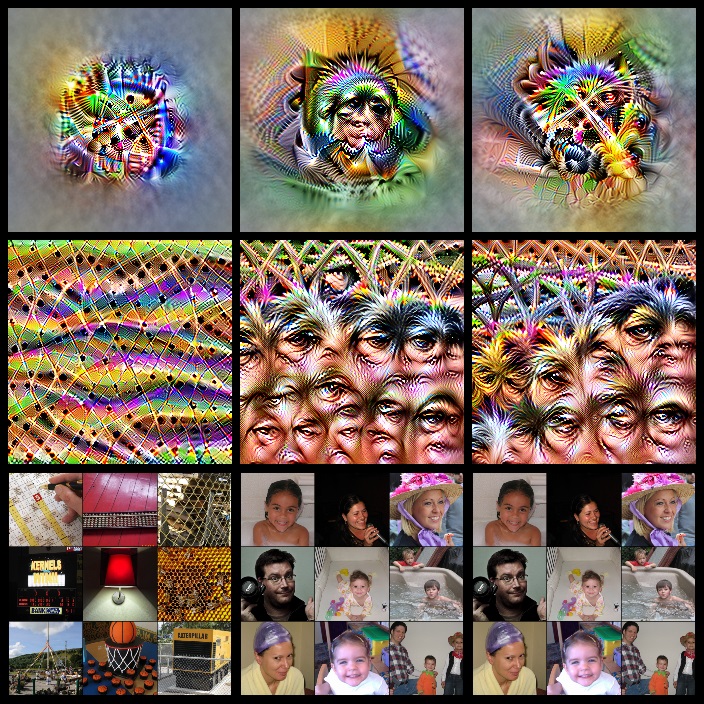}
        220
        \vspace{0.15\linewidth}
    \end{subfigure}
    \begin{subfigure}{.12\linewidth}
        \centering
        \includegraphics[width=\linewidth]{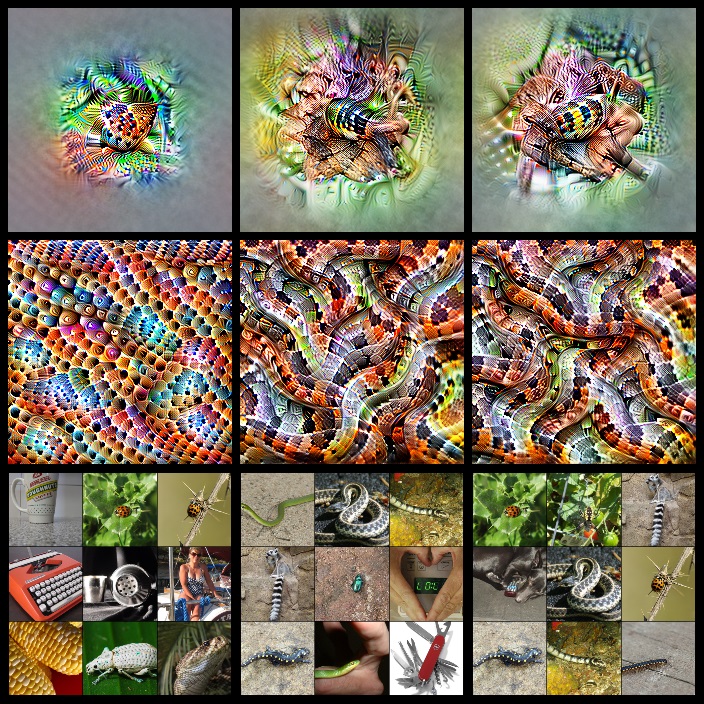}
        226
        \vspace{0.15\linewidth}
    \end{subfigure}
    \begin{subfigure}{.12\linewidth}
        \centering
        \includegraphics[width=\linewidth]{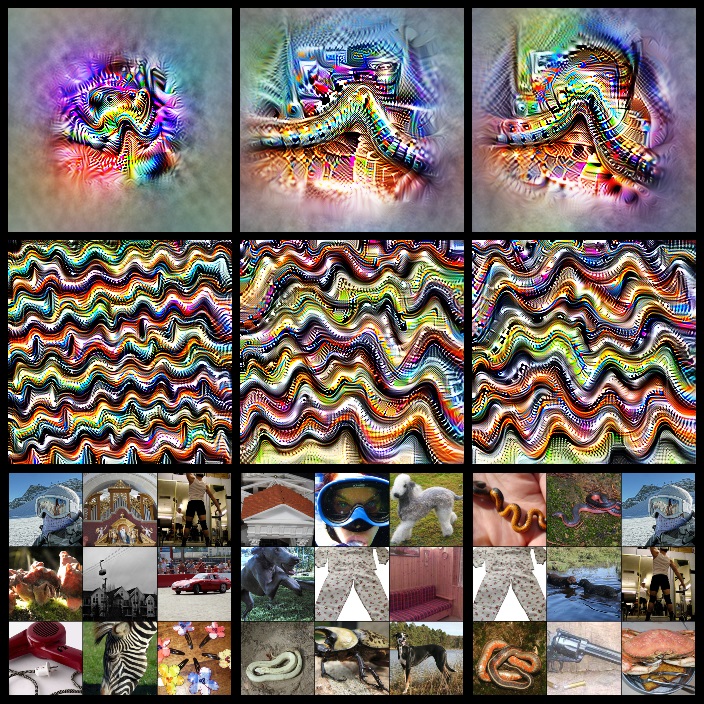}
        250
        \vspace{0.15\linewidth}
    \end{subfigure}
    
    \caption{Grids of maximally exciting images for all remaining scale invariant criteria-passing channels in block 3.1.  All channels are zero-indexed.}
    \label{fig:A2}
\end{figure*}

We present the grids of all criteria-passing channels throughout ResNet18 in \cref{fig:A1} (blocks 1.1, 2.0, and 2.1) and \cref{fig:A2} (block 3.1).

